\documentclass{article}

\usepackage{microtype}
\usepackage{graphicx}
\usepackage{subcaption}
\usepackage{booktabs} % for professional tables

\usepackage{hyperref}

\usepackage[accepted]{icml2026}

\usepackage{amsmath}
\usepackage{amssymb}
\usepackage{mathtools}
\usepackage{amsthm}

\usepackage[capitalize,noabbrev]{cleveref}

\theoremstyle{plain}

\theoremstyle{definition}

\theoremstyle{remark}

\usepackage[textsize=tiny]{todonotes}

\icmltitlerunning{MMBench-Live: A Continuously Evolving Benchmark for Multimodal Models}

\begin{document}

\twocolumn[
  \icmltitle{MMBench-Live: A Continuously Evolving Benchmark for Multimodal Models}



  \icmlsetsymbol{equal}{*}

    \begin{icmlauthorlist}
      \icmlauthor{Yuanzhi Liu}{bupt}
      \icmlauthor{Shousheng Zhao}{bupt}
      \icmlauthor{Bo Zhou}{bupt}
      \icmlauthor{Kongming Liang}{bupt}
      \icmlauthor{Zhanyu Ma}{bupt}
    \end{icmlauthorlist}
    
    \icmlaffiliation{bupt}{Beijing University of Posts and Telecommunications, Beijing, China}
    
    \icmlcorrespondingauthor{Kongming Liang}{liangkongming@bupt.edu.cn}

  \icmlkeywords{Machine Learning, ICML}

  \vskip 0.3in
]



\printAffiliationsAndNotice{}  % no special notice (required even if empty)

\begin{abstract}
Evaluation benchmarks are essential for assessing vision--language models (VLMs), but most multimodal benchmarks are static, making them vulnerable to temporal staleness, data contamination, and costly maintenance. We present MMBench-Live, a continuously evolving multimodal benchmark built by a multi-agent-driven automated pipeline. Our framework treats benchmark evolution as task-guided dataset construction, integrating structured benchmark specification, feedback-controlled real-time data acquisition, and verifiable QA generation with executable reasoning. To maintain cross-version comparability, we introduce a distribution-consistent update strategy that extracts task-related visual patterns from the original benchmark to guide data collection and filtering. Instantiated from MMBench, MMBench-Live contains 5.9K newly generated evaluation instances with a high answer correctness rate, while each update costs about \$30 and takes 1--2 hours. Extensive evaluations show that MMBench-Live preserves stable model rankings, maintains semantic alignment with the original benchmark, and exhibits weaker contamination-related memorization signals, suggesting a practical and scalable paradigm for sustainable multimodal benchmark evolution. The project is available at \url{https://github.com/PRIS-CV/MMBench-Live}.

\end{abstract}

\section{Introduction}
Most existing vision-language evaluation benchmarks~\cite{lu2022learn,fu2023mme,Li_2024_CVPR,liu2024mmbench} adopt a static evaluation paradigm, implicitly assuming that a fixed test set can serve as a long-term proxy for model capability. This assumption becomes increasingly fragile as modern vision-language models (VLMs) scale rapidly and are trained on ever-growing web-scale corpora. Static benchmarks are vulnerable to temporal staleness, data contamination~\citep{Touvron2023LLaMAOA,chen2024are,song2025both}, and high maintenance costs, making frequent updates impractical at scale. However, dynamically updating a benchmark is not simply a matter of replacing old samples with new ones: a reliable live benchmark must introduce temporally fresh instances while preserving the original task semantics, capability coverage, data characteristics, and cross-version comparability.

Prior work has explored several strategies for updating multimodal evaluation datasets. Self-evolving methods based on visual perturbations or linguistic rewrites~\citep{yang2025dynamic,zhang2026kbedmedynamicmultimodalevaluation} may suffer from semantic drift and degraded comparability, while generation-based methods~\citep{wang2026sdevalsafetydynamicevaluation,zhang2025dysca,wen2025spot} can introduce distributional gaps under complex semantic constraints. Real-world data updating~\citep{jiang2025mac,shabtay2025livexiv,jain2025livecodebench,white2025livebench} is a promising direction because it naturally improves temporal freshness and reduces direct reuse of static benchmark samples. Nevertheless, existing pipelines are often tied to fixed data sources, predefined task formats, or task-specific construction procedures. Consequently, continuously incorporating fresh real-world data across diverse multimodal tasks while preserving semantic coherence, distributional alignment, and evaluation stability remains challenging.

To address this challenge, we propose MMBench-Live, a continuously evolving multimodal benchmark built through a multi-agent-driven automated pipeline. Our framework models benchmark evolution as task-guided dataset construction. Starting from the original benchmark, it converts evaluation objectives, task hierarchy, and atomic tasks into structured benchmark descriptions, which serve as the semantic basis for data acquisition and QA generation. To preserve cross-version comparability, MMBench-Live identifies task-related visual patterns from the original benchmark and uses them to guide feedback-controlled real-time data acquisition, enabling the pipeline to refine poorly aligned retrieval queries and filter task-irrelevant candidates. For each collected image, the QA generation stage constructs a question, an answer, and an executable solution plan, which is verified through tool-supported reasoning to improve the reliability of automatically generated evaluation instances.

We instantiate this framework by systematically updating MMBench~\citep{liu2024mmbench}. The resulting MMBench-Live contains 5.9K newly generated evaluation instances, achieves a manual answer correctness rate of 96.06\%, and completes each update within approximately 1--2 hours at a cost of about \$30. Experiments on representative open-source VLMs show that MMBench-Live maintains stable cross-version model rankings, preserves the core semantic and distributional characteristics of the original benchmark, and exhibits weaker PaCoST-based~\cite{zhang-etal-2024-pacost} memorization signals than MMBench. These results demonstrate that MMBench-Live provides a practical, scalable, and low-cost paradigm for sustainable multimodal benchmark evolution.

The main contributions can be summarized as follows:
\begin{itemize}
    \item We introduce MMBench-Live, a continuously evolving multimodal benchmark constructed through a multi-agent-driven automated pipeline, enabling scalable and cost-efficient benchmark updates with temporally fresh real-world data.
    
    \item We propose a distribution-consistent benchmark updating strategy that combines task-related visual pattern identification with feedback-controlled data acquisition, preserving task semantics and visual characteristics across benchmark versions.
    
    \item We conduct systematic evaluations on MMBench-Live, showing high QA correctness, low construction cost, stable cross-version consistency, distributional alignment, and weaker contamination-related memorization signals.
\end{itemize}

\section{Related Work}

\subsection{Data Contamination in Model Evaluation}

Data contamination has become a critical issue in evaluating large language models (LLMs) and vision-language models (LVLMs), as overlap between benchmarks and pretraining corpora can substantially inflate reported performance. Prior studies show that such leakage is widespread even in commonly used benchmarks; for example, the LLaMA-2 report finds that over 16\% of MMLU samples are contaminated, with a non-trivial fraction exhibiting severe leakage~\cite{touvron2023llama}. Controlled experiments further demonstrate that even limited contamination can cause significant score inflation, disproportionately favoring larger models~\cite{kocyigit2025overestimation}. Position papers argue that these effects compromise the validity of model comparisons and lead to misleading conclusions about model capabilities~\cite{sainz-etal-2023-nlp}. The issue is exacerbated in multimodal evaluation, where leakage in either textual or visual modalities can artificially boost performance on vision-language benchmarks~\cite{song-etal-2025-text}. Together, these findings underscore the vulnerability of static benchmarks as pretraining corpora scale, motivating recent efforts toward automated and dynamic evaluation datasets designed to mitigate contamination~\cite{wu-etal-2025-antileakbench}.

\subsection{Automatic Dataset Construction}

Automatic dataset construction has gained increasing attention as manual curation and annotation limit the scalability of training and evaluation corpora. A major research direction frames dataset construction as automated instruction synthesis, where large language models generate and filter new instances from small seed sets, including reasoning-oriented and curriculum-based variants~\cite{wang-etal-2023-self-instruct,bu-etal-2025-enhanced,xu2024wizardlm}. These principles have been extended to multimodal settings, enabling large-scale vision-language instruction data synthesis with minimal human annotation, as exemplified by LLaVA~\cite{liu2023visual_instruction_tuning}. Complementary approaches further incorporate programmatic structures and external artifacts—such as tool specifications, executable APIs, and web-scale data collection with lightweight filtering—to improve reliability and reduce manual effort~\cite{guo2025api,xu2023toolbench,liu2024adc}. However, most existing methods rely on post-hoc filtering rather than systematic validation, motivating closed-loop construction pipelines that explicitly enforce data correctness and evaluation relevance.

\begin{figure*}[t]
  \centering
  \includegraphics[width=0.95\textwidth]{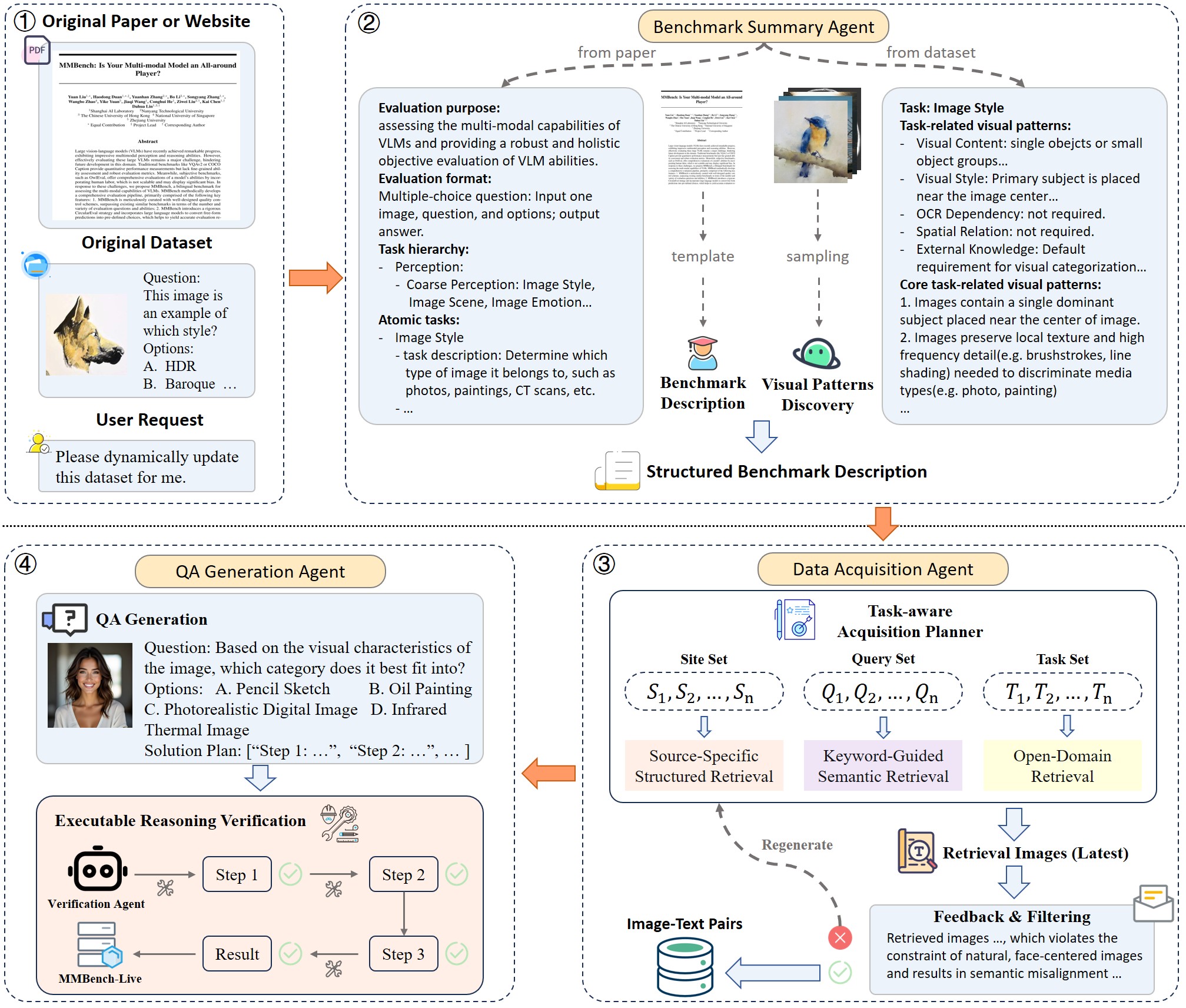}
  \caption{Overview of the MMBench-Live framework. MMBench-Live is constructed by converting the original MMBench into structured descriptions, acquiring new data in a task-aware manner, and generating executable and verified QA pairs for evaluation.}
  \label{fig:pipeline}
\end{figure*}

\section{MMBench-Live}
We construct MMBench-Live by converting the original MMBench into structured descriptions, acquiring new data in a task-aware manner, and generating executable and verified QA pairs for evaluation. Therefore, it can maintain (i) capability coverage consistency, (ii) distributional alignment with the original benchmark, and (iii) evaluation discriminativeness over time.

\subsection{Benchmark Summary}
To enable controlled and stable benchmark updates, we convert MMBench into a structured description, providing a unified representation for subsequent data acquisition and instance construction.
Inspired by the prior works~\cite{reuel2024betterbench,sokol2025benchmarkcards}, we define a structured benchmark template for MMBench, including four components:

(1) \textbf{Evaluation purpose.} The evaluation purpose specifies the target capabilities that the benchmark is designed to assess. It provides a high-level semantic characterization of what competencies are being evaluated, such as visual perception, cross-modal reasoning, spatial understanding, or compositional inference.

(2) \textbf{Evaluation format.} The evaluation format specifies the structural form of an evaluation instance, including the modalities involved, the input–output interface, and the expected response type. An evaluation instance can be a single image paired with a multiple-choice question, an open-ended textual answer, or a structured output.

(3) \textbf{Task hierarchy.} The task hierarchy organizes the benchmark into a multi-level structure of tasks with various granularity. At the top level, the hierarchy captures coarse-grained evaluation tasks, which are progressively decomposed into finer-grained sub-tasks that correspond to specific and well-scoped capability requirements. For example, in MMBench, high-level \textit{perception} tasks are further refined into fine-grained tasks such as \textit{OCR} and \textit{attribute recognition}. 

(4) \textbf{Atomic tasks.} Atomic tasks correspond to the leaf-level units in the task hierarchy and define the smallest operational evaluation units. For each atomic task, we also define a structured task description template consisting of three components: \emph{Task purpose}, \emph{Data sources and construction procedures} and \emph{Task characterization}.

Task characterization is decomposed into textual and visual components. The textual aspect is represented by QA examples, which clarify the intended evaluation semantics. The visual aspect is captured by task-related visual patterns, defined as recurrent visual or semantic attributes that occur frequently within an atomic task but are not explicitly specified in the task definition, implicitly constraining valid evaluation instances.

To identify these patterns, we adopt a data-driven procedure summarized in Algorithm~\ref{alg:task_visual_patterns}. For each atomic task $t$, we sample a subset of original instances and analyze them from five semantic perspectives inspired by MMBench: visual content, visual style, OCR dependency, spatial relations, and external knowledge dependency. A VLM extracts recurring patterns $C_t$ under each perspective, which are then consolidated by an LLM to merge similar patterns and remove redundancy, yielding a compact set of task-related visual patterns. These patterns guide subsequent data acquisition and filtering, ensuring consistency with the visual characteristics of the original evaluation data.

\begin{algorithm}[t]
\caption{Task-Related Visual Pattern Identification}
\label{alg:task_visual_patterns}
\begin{algorithmic}[1]
\STATE \textbf{Input:} evaluation instances $E_t$ for atomic task $t$; sampling ratio $\rho$; analysis perspectives $P$
\STATE \textbf{Output:} task-related visual pattern set $B_t$

\STATE $E'_t \leftarrow \mathrm{Sample}(E_t, \rho)$
\STATE $C_t \leftarrow \emptyset$
\FOR{$p \in P$}
    \STATE $C_p \leftarrow \mathrm{VLM\_Extract}(E'_t, p)$
    \STATE $C_t \leftarrow C_t \cup C_p$
\ENDFOR
\STATE $B_t \leftarrow \mathrm{LLM\_Consolidate}(C_t)$
\STATE \textbf{return} $B_t$
\end{algorithmic}
\end{algorithm}

\subsection{Task-aware Data Acquisition}
This stage collects new evaluation data by converting each atomic task into an executable acquisition procedure. Rather than issuing generic search queries, the acquisition process is constrained by the task intent, required visual evidence, and available acquisition context. It consists of three components: an acquisition planner, an acquisition executor, and a feedback controller.

\textbf{Acquisition Planner.}
For each atomic task $t$, the acquisition planner analyzes the structured task description $S_t$ and produces an acquisition plan:
\begin{equation}
P_t = \langle F_t,\; A_t,\; Q_t \rangle .
\end{equation}
Here, $F_t$ denotes the observable fields that must be preserved with each collected image, such as titles, surrounding textual context, source metadata, or timestamps. $A_t$ denotes the acquisition strategy, and $Q_t$ denotes the generated query set when query-based retrieval is required.

\textbf{Acquisition Executor.}
The Acquisition executor performs data collection according to the plan generated by acquisition planner. It supports three executable retrieval modes:

(1) Source-specific Structured Retrieval.
This mode conducts structured crawling over designated websites specified in the acquisition plan. Relevant image links and associated textual fields are extracted from the crawler outputs and retained. This mode is primarily used when authoritative or task-specific data sources are explicitly defined.

(2) Keyword-Guided Semantic Retrieval.
This mode submits a set of query keywords to the Google Image API to retrieve candidate images. To maintain temporal relevance and reduce potential contamination, retrieval is restricted to images uploaded within a one-year time window, and associated metadata are retained.

(3) Open-Domain Retrieval.
This mode continuously samples recently uploaded public images from open-domain content streams via the Flickr API, together with their associated metadata. This mode does not rely on predefined sources or explicit semantic constraints, enabling broad coverage of diverse and naturally occurring visual content.

% \textbf{Feedback Controller.}
% To maintain semantic alignment during data acquisition, we introduce a lightweight feedback controller that performs task-conditioned validation and filtering over collected samples. Unlike one-shot acquisition, the controller operates on-the-fly during data collection and does not require updating any model parameters.

% During acquisition, candidate samples are collected in batches, and a small subset is periodically evaluated using a multimodal judge model. Given a candidate image $x$ and its associated textual context $t$, the judge outputs a binary consistency decision, a confidence score, and a brief natural-language diagnostic signal with respect to the task purpose $G_t$ and the task-related visual patterns $B_t$:
% \begin{equation}
% (y, s, r) = J(x, t \mid G_t, B_t), \quad y \in \{0,1\},
% \end{equation}
% where $J(\cdot)$ denotes a multimodal judge model, $s$ represents the confidence of the decision, and $r$ is a short diagnostic message explaining the reason for mismatch when applicable.
% Samples predicted as misaligned with high confidence are filtered out. The diagnostic signal $r$ is used to adjust subsequent acquisition behavior.

\textbf{Feedback Controller.}
During data acquisition, retrieval queries may return visually plausible but task-irrelevant images, especially when the task involves fine-grained visual cues, implicit constraints, or subjective concepts. To maintain semantic alignment, we introduce a lightweight feedback controller that verifies retrieved candidates and refines poorly aligned queries during retrieval. The controller is guided by the task purpose $G_t$ and the task-related visual patterns $B_t$, which jointly specify the intended semantics and expected visual evidence of valid samples for task $t$.

For each retrieved candidate image $x_i$ with its associated textual context $c_i$, we instantiate a prompt-guided multimodal judge $J$ to evaluate whether the candidate is visually consistent with the task requirements:
\begin{equation}
(a_i, r_i) = J(x_i, c_i \mid G_t, B_t),
\end{equation}
where $J$ denotes a multimodal judging model guided by task-specific prompts. Given the candidate image $x_i$, its textual context $c_i$, the task purpose $G_t$, and the task-related visual patterns $B_t$, the prompt instructs $J$ to check task consistency based on observable visual evidence and to output a binary acceptance decision $a_i \in \{0,1\}$ together with one to three concise visual rationales $r_i$. These rationales are grounded in observable image evidence, such as object presence, scene type, visual attributes, action states, spatial relations, or other task-specific visual cues.

For each retrieval query, the controller examines the top-$K$ returned candidates. In our implementation, we set $K=10$. The acceptance rate of the current query is computed as
\begin{equation}
\rho_q =
\frac{1}{K}
\sum_{i=1}^{K} a_i .
\end{equation}

If $\rho_{q_m} \geq \tau$, the current query is considered sufficiently aligned and is retained for subsequent acquisition. Otherwise, the controller treats the rejected candidates as diagnostic examples. Their visual rationales are aggregated into a compact feedback signal $r$, which conditions the query update:
\begin{equation}
q_{m+1}
=
U(q_m \mid G_t, B_t, r),
\quad \text{if } \rho_{q_m} < \tau .
\end{equation}
The update operator $U$ is instantiated with a large language model. In this way, the updated query preserves the task intent while reducing the visual deviations observed in the previous retrieval round.

As summarized in Algorithm~\ref{alg:feedback_controller}, this feedback process is repeated for at most three iterations. Once the query satisfies the acceptance criterion or reaches the maximum number of iterations, the controller stops updating the retrieval query. The final query is then used for subsequent candidate collection, while the judge is applied only as a candidate-level filter. This design allows the acquisition process to correct poorly aligned search queries early, while keeping the overall pipeline lightweight. It improves the semantic quality of collected samples without human annotation, model fine-tuning, or updates to the underlying judge model.

\begin{algorithm}[t]
\caption{Feedback-controlled data acquisition}
\label{alg:feedback_controller}
\begin{algorithmic}[1]
\REQUIRE $G_t$, $B_t$, $q_0$, $J$, $K=10$, $\tau=0.7$, $M=3$
\ENSURE Filtered candidate set $\mathcal{D}^{+}$
\FOR{$m=0$ to $M-1$}
    \STATE $\mathcal{D}_m \leftarrow \operatorname{Retrieve}(q_m, K)$
    \STATE $\{(a_i,r_i)\}_{i=1}^{K} \leftarrow 
    \{J(x_i,c_i \mid G_t,B_t)\}_{(x_i,c_i)\in \mathcal{D}_m}$
    \STATE $\rho_{q_m} \leftarrow \frac{1}{K}\sum_{i=1}^{K} a_i$
    \IF{$\rho_{q_m} \geq \tau$}
        \STATE $q^\star \leftarrow q_m$
        \STATE \textbf{break}
    \ENDIF
    \STATE $r \leftarrow \operatorname{Agg}(\{r_i \mid a_i=0\})$
    \STATE $q_{m+1} \leftarrow U(q_m \mid G_t,B_t,r)$
\ENDFOR
\IF{$q^\star$ is undefined}
    \STATE $q^\star \leftarrow q_m$
\ENDIF
\STATE $\mathcal{D} \leftarrow \operatorname{Retrieve}(q^\star)$
\STATE $\mathcal{D}^{+} \leftarrow \{(x,c)\in \mathcal{D} \mid J(x,c \mid G_t,B_t)=1\}$
\STATE \textbf{return} $\mathcal{D}^{+}$
\end{algorithmic}
\end{algorithm}

\subsection{QA Generation and Verification}
This stage constructs generate QA pairs from image data under a given task specification and verifies them through executable reasoning, ensuring that each evaluation instance is semantically aligned with the target task and admits a deterministic validation procedure.

\textbf{QA Generation.}
Given an atomic task $t$ and an image $x$, the QA generator constructs a task-consistent and verifiable QA instance grounded in the image. To reduce systematic generation bias and improve generation quality, multiple multimodal large models are used to produce candidate QA pairs, and the best candidate is selected through model-based comparison and rule-based validity checks.

For each image, the generated instance is represented as a structured triplet:
\begin{equation}
z = (q, a, \pi),
\end{equation}
where $q$ is the evaluation question, $a$ is the corresponding answer, and $\pi$ is an executable solution plan specifying how $a$ can be derived from the image. The solution plan provides an explicit reasoning procedure for subsequent verification through tool execution.

Each generated triplet is required to satisfy three criteria: consistency with the atomic task definition, explicit reliance on visual evidence from the image, and executability of the solution plan. This design turns QA generation from unconstrained question writing into structured instance construction, where the generated question, answer, and solution plan are jointly prepared for automatic validation.

\textbf{Executable Reasoning Verification}
To address visual hallucinations and enhance the reliability of automatically generated QA pairs, we explicitly decouple correctness verification from direct vision-based model judgment and instead employ a text-driven executable reasoning process for validation.

Given a synthesized instance $(q, a, \pi)$, the solution plan $\pi$ is treated as a structured and interpretable sequence of executable reasoning operations. Specifically, we represent the plan as
\begin{equation}
\pi = (o_1, o_2, \ldots, o_K),
\end{equation}
where $o_k$ denotes the executable operation at step $k$.

Importantly, the verification controller itself is vision-blind: it does not directly access the image but performs reasoning solely over the textual intermediate results generated by tool execution.
Starting from the input image $x$, tools are executed sequentially according to the generated operation sequence to produce textual intermediate outputs:
\begin{equation}
z_k = T(o_k \mid x, z_{<k}), \quad k=1,\dots,K,
\end{equation}
where $T$ denotes the tool execution process, and $z_{<k}$ represents previously obtained textual intermediate outputs.

The final verification outcome $\hat{a}$ is then predicted by a vision-blind language-model verifier conditioned only on the question and the tool-derived textual observations:
\begin{equation}
\hat{a}
=
\arg\max_y P\big(y \mid q, z_{1:K}\big).
\end{equation}

The synthesized answer $a$ and the corresponding QA pair are accepted only if it matches the verified outcome $\hat{a}$,
\begin{equation}
\mathbb{I}(q,a,\pi) = \mathbb{I}\big[a = \hat{a}\big].
\end{equation}

Overall, this design constrains the verification process to explicit reasoning chains, thereby reducing reliance on implicit visual guessing and mitigating hallucinations.

\begin{table*}[t]
\centering
\caption{Results on MMBench-Live (L-2 abilities). Abbreviations adopted: LR for Logical Reasoning; AR for Attribute Reasoning; RR for Relation Reasoning; FP-C for Fine-grained Perception (Cross Instance); FP-S for Fine-grained Perception (Single Instance); CP for Coarse Perception.}
\label{tab:dynamic_mmbench_l2}
\begin{tabular}{@{}lccccccc@{}}
\toprule
Model & Overall & CP & FP-S & FP-C & AR & LR & RR \\
\midrule
DeepSeek-VL-7B-Chat~\cite{lu2024deepseekvl}
& 83.04\% & \textbf{97.37\%} & \textbf{75.37\%} & 68.22\% & \textbf{94.00\%} & 67.17\% & 85.60\% \\
InstructBLIP-Vicuna-7B~\cite{dai2023instructblip}
& 73.85\% & 92.05\% & 60.14\% & 55.66\% & 91.73\% & 56.95\% & 73.44\% \\
LLaVA-v1.5-7B~\cite{liu2023llava}
& 74.83\% & 91.56\% & 66.14\% & 56.80\% & 86.98\% & 59.13\% & 76.98\% \\
mPLUG-Owl2-7B~\cite{10657415}
& 77.18\% & 93.36\% & 66.49\% & 60.33\% & 90.94\% & 60.64\% & 80.52\% \\
Qwen3-VL-8B-Instruct~\cite{qwen3technicalreport}
& 80.41\% & 80.36\% & 72.76\% & 70.72\% & 88.11\% & 72.36\% & 84.42\% \\
Qwen2.5-VL-7B-Instruct~\cite{bai2025qwen25vltechnicalreport}
& \textbf{83.10\%} & 81.54\% & \textbf 75.11\% & \textbf{77.47\%} & 88.45\% & \textbf{80.23\%} & \textbf{85.71\%} \\
\bottomrule
\end{tabular}
\end{table*}

\begin{table*}[t]
\centering
\caption{Cross-version correlation of model performance between the original and updated benchmarks.}
\label{tab:cross_version_correlation}
\begin{tabular}{lcccc}
\toprule
Granularity &
Pearson $r$ & Pearson $p$ &
Spearman $\rho$ & Spearman $p$ \\
\midrule
Task-Model Level &
0.5200 & $1.15\times10^{-9}$ &
0.4254 & $1.27\times10^{-6}$ \\
Model-Averaged Level &
0.8876 & 0.01823 &
0.8286 & 0.04156 \\
\bottomrule
\end{tabular}%
\end{table*}

% \begin{table*}[t]
% \centering
% \caption{Cross-version correlation of model performance between the original and updated benchmarks for different benchmark versions.}
% \label{tab:cross_version_correlation_versions}
% \begin{tabular}{llcccc}
% \toprule
% Version & Granularity &
% Pearson $r$ & Pearson $p$ &
% Spearman $\rho$ & Spearman $p$ \\
% \midrule
% v1 & Task-Model Level &
% 0.5200 & $1.15\times10^{-9}$ &
% 0.4254 & $1.27\times10^{-6}$ \\
% v1 & Model-Averaged Level &
% 0.8876 & 0.01823 &
% 0.8286 & 0.04156 \\
% \midrule
% v2 & Task-Model Level &
% 0.6039 & $2.835\times10^{-13}$ &
% 0.5568 & $3.995\times10^{-11}$ \\
% v2 & Model-Averaged Level &
% 0.9534 & 0.003203 &
% 0.9429 & 0.004805 \\
% \bottomrule
% \end{tabular}%
% \end{table*}

\section{Experiment}
We evaluate the proposed framework through extensive experiments on MMBench-Live. Our analysis covers model performance, dataset quality, construction cost, distributional alignment, cross-version consistency, and data contamination, together with ablation studies examining the role of task constraints, semantic feedback, and underlying foundation models.

\subsection{Evaluation Setup}
\textbf{Size of Dataset.} Taking the MMBench~\cite{liu2024mmbench} dev set (4K) as the reference baseline, we apply our automated framework to construct MMBench-Live, which results in 5.9K newly generated evaluation QA pairs.

\textbf{Model Selected.} In the QA Generation stage, we employ a small pool of generation models, consisting of OpenAI GPT-5 Mini~\cite{openai2025gpt5} and Gemini-3-Flash-Preview~\cite{gemmateam2025gemma3technicalreport}, to construct QA pairs. All agent components are uniformly instantiated with OpenAI GPT-5 Mini as the underlying proxy model.

For executable reasoning verification, VisionReasoner~\cite{liu2026visionreasoner} is used for object detection and segmentation, Depth Anything V2~\cite{yang2024depth} is used for monocular depth estimation, and a fine-tuned LLaVA~\cite{liu2023llava} model is used for visual attribute recognition. For text-related perception, Qwen2.5-VL~\cite{bai2025qwen25vltechnicalreport} and GOT-OCR-2-HF~\cite{wei2024generalocrtheoryocr20} are used for OCR. Finally, Qwen3~\cite{yang2025qwen3technicalreport} is adopted as the text-only verifier, which aggregates the intermediate tool outputs and performs the final reasoning step.

For evaluation, we consider a diverse set of representative VLMs, including DeepSeek-VL-7B-Chat~\cite{lu2024deepseekvl}, InstructBLIP-Vicuna-7B~\cite{dai2023instructblip}, mPLUG-Owl2-7B~\cite{10657415}, LLaVA-v1.5-7B~\cite{liu2023llava}, Qwen3-VL-8B-Instruct~\cite{qwen3technicalreport}, and Qwen2.5-VL-7B-Instruct~\cite{bai2025qwen25vltechnicalreport}. To guarantee fair and deterministic comparisons, the generation temperature of all evaluated models is fixed to zero during inference.

\subsection{Evaluation Results on MMBench-Live}
Table \ref{tab:dynamic_mmbench_l2} summarizes the performance of different models in MMBench-Live. Overall, Qwen2.5-VL achieves the best aggregate performance, slightly outperforming DeepSeek-VL. In contrast, earlier models obtain substantially lower scores, indicating limited its generalization capability to continuously evolving benchmarks. A clear performance divergence is observed across evaluation dimensions. DeepSeek-VL excels in perception-oriented tasks, achieving the highest scores on coarse perception and attribute recognition, reflecting strong coarse-level visual understanding and attribute recognition ability. In comparison, Qwen2.5-VL consistently outperforms other models in fine-grained perception and reasoning-intensive dimensions, highlighting its advantage in detailed visual understanding and logical reasoning.

We also observe that Qwen2.5-VL outperforms the more recent Qwen3-VL across multiple evaluation dimensions. A similar performance reversal is also observed on MMBench. The performance degradation of Qwen3-VL is primarily concentrated in cross-instance fine-grained comparison and spatial reasoning tasks, where the model more frequently exhibits misaligned comparison targets and incorrect reference-frame binding.

\begin{figure*}[t]
  \centering
  \includegraphics[width=0.95\textwidth]{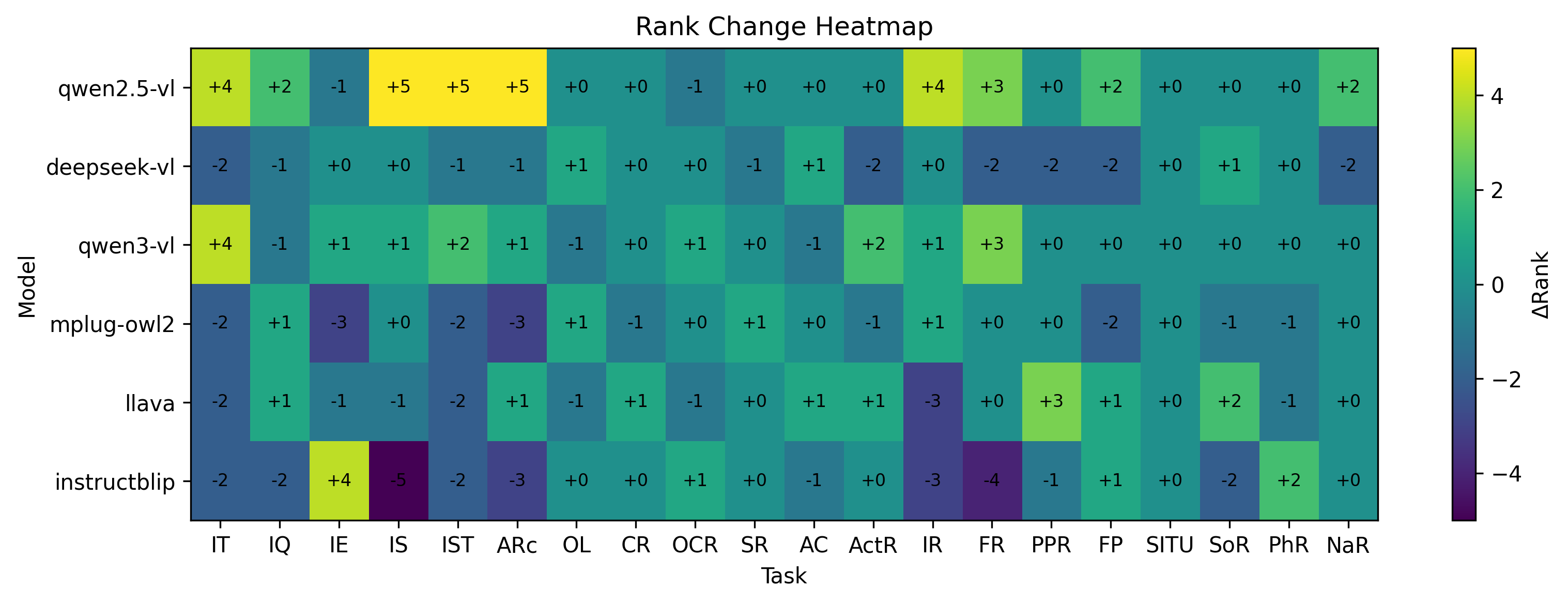}
  \caption{Rank change heatmap between MMBench and MMBench-Live. Each cell shows the rank difference $\Delta \mathrm{Rank} = \mathrm{Rank}{\mathrm{Live}} - \mathrm{Rank}{\mathrm{MMBench}}$ for a given model–task pair. All tasks are denoted using abbreviated names and full task names are provided in the appendix~\ref{sec:appendix:abbrev}.}
  \label{fig:rank_heatmap}
\end{figure*}

As shown in Figure~\ref{fig:rank_heatmap}, the transition from MMBench to MMBench-Live does not induce global ranking instability. Instead, ranking variations are largely driven by a small subset of highly discriminative sub-tasks, indicating that dynamic benchmark updates can surface fine-grained capability differences while preserving overall evaluation stability.

\begin{table*}[t]
\centering
\caption{Overall data contamination comparison between MMBench and MMBench-Live.}
\label{tab:contamination_comparison}
\begin{tabular}{lcccccc}
\toprule
Benchmark & Mean $\Delta$ & Std $\Delta$ & $t$-stat & $p$ (one-sided) & Cohen's $d$ \\
\midrule
MMBench~\cite{liu2024mmbench} & 0.0118 & 0.1220 & 6.38 & $9.5\times10^{-11}$ & 0.097 \\
MMBench-Live & \textbf{0.0060} & \textbf{0.1016} & 4.54 & $2.9\times10^{-6}$  & \textbf{0.059} \\
\bottomrule
\end{tabular}%
\end{table*}

% \begin{table*}[t]
% \centering
% \caption{Overall data contamination comparison between MMBench and MMBench-Live.}
% \label{tab:contamination_comparison}
% \begin{tabular}{lcccccc}
% \toprule
% Benchmark & Mean $\Delta$ & Std $\Delta$ & Cohen's $d$ \\
% \midrule
% MMBench~\cite{liu2024mmbench} & 0.0118 & 0.1220 & 0.097 \\
% MMBench-Live & \textbf{0.0060} & \textbf{0.1016} & \textbf{0.059} \\
% \bottomrule
% \end{tabular}%
% \end{table*}

% \begin{table*}[t]
% \centering
% \caption{Cross-version consistency of the Qwen-based MMBench-Live.}
% \label{tab:qwen_cross_version_consistency}
% \begin{tabular}{ccc|ccc}
% \toprule
% \textbf{Granularity} & \textbf{Pearson $r$} & \textbf{Spearman $\rho$}
% & \textbf{Granularity} & \textbf{Pearson $r$} & \textbf{Spearman $\rho$} \\
% \midrule
% Task-Model 
% & 0.4574 
% & 0.4408 
% & Model-Averaged 
% & 0.9084 
% & 0.8286 \\
% \bottomrule
% \end{tabular}
% \end{table*}

% \begin{table*}[t]
% \centering
% \caption{Overall data contamination comparison between MMBench, MMBench-Live, and additional methods from VLB~\cite{yang2025dynamic}.}
% \label{tab:contamination_comparison_updated}
% \begin{tabular}{lcccc}
% \toprule
% Method & Mean $\Delta$ & Std $\Delta$ & Cohen's $d$ \\
% \midrule
% MMBench~\cite{liu2024mmbench} & 0.012 & 0.122 & 0.097 \\
% Add~\cite{yang2025dynamic} & 0.042 & 0.187 & 0.224 \\
% Remove~\cite{yang2025dynamic} & 0.038 & 0.200 & 0.191 \\
% Outpainting~\cite{yang2025dynamic} & 0.042 & 0.205 & 0.205 \\
% MMBench-Live & \textbf{0.006} & \textbf{0.102} & \textbf{0.059} \\
% \bottomrule
% \end{tabular}%
% \end{table*}

\subsection{Dataset Quality and Construction Cost}
\textbf{QA Correctness.}
To evaluate the correctness of automatically constructed QA pairs, we conduct a manual verification study. For each task, 10\% of the images are randomly sampled, and their corresponding QA pairs are manually verified. This procedure is repeated three times with independent samples to reduce variance. Across all tasks and trials, the QA pairs achieve an average answer correctness rate of 96.06\%, demonstrating high annotation reliability. This statistic measures the correctness of the final answers only, rather than the validity of the executable solution plans. For failure cases, we provide a representative case analysis in Appendix~\ref{app:failure_analysis}.

\textbf{Construction Cost.} We evaluate the construction cost from both monetary and time perspectives. Prior studies~\cite{tang2024mtvqa,zhang2025qapyramid,Chen_2025_ICCV} on manual dataset construction report that, for a dataset at the scale of the 6K-sample MMBench-Live, manual pipelines typically incur costs on the order of thousands to tens of thousands of dollars, with turnaround times of approximately one to two months. In contrast, our automated pipeline completes a benchmark update within approximately 1–2 hours, with a total monetary cost of about \$30, where QA generation accounts for the majority of the expense (around 60\%). Overall, the proposed approach reduces both construction cost and turnaround time by multiple orders of magnitude, enabling frequent and sustainable benchmark updates.

\begin{table*}[t]
\centering
\caption{Distribution alignment between the dynamically updated dataset and the original dataset, where A denotes data collected by the dynamic update pipeline (with or without task-related visual patterns and semantic feedback), and B denotes the MMBench dataset.}
\label{tab:dist_alignment}
\small
\setlength{\tabcolsep}{6pt}
\begin{tabular}{lcccc}
\toprule
\textbf{Updated Dataset} 
& \textbf{FID} $\downarrow$ \textbf{(mean$\pm$std)}
& \textbf{PCA-KL(A$\parallel$B)} $\downarrow$
& \textbf{PCA-KL(B$\parallel$A)} $\downarrow$
& $\boldsymbol{\Delta}$\textbf{KL(A$\parallel$B $-$ B$\parallel$A)} \\
\midrule
dataset (w/ visual patterns + feedback)
& \textbf{0.1831 $\pm$ 0.0007} 
& 109.321 
& \textbf{95.514}
& \textbf{+13.807} \\

dataset (w/o visual patterns + feedback)
& 0.2035 $\pm$ 0.0012 
& \textbf{108.261} 
& 142.969 
& \textbf{-34.708} \\
\bottomrule
\end{tabular}
\end{table*}

\begin{table*}[t]
\centering
\caption{Results on MMBench-Live (L-2 abilities) on the Qwen-based version.
Abbreviations adopted:
LR for Logical Reasoning; AR for Attribute Reasoning; RR for Relation Reasoning;
FP-C for Fine-grained Perception (Cross Instance); FP-S for Fine-grained Perception (Single Instance);
CP for Coarse Perception.}
\label{tab:dynamic_mmbench_l2_qwen}
\begin{tabular}{@{}lccccccc@{}}
\toprule
Model & Overall & CP & FP-S & FP-C & AR & LR & RR \\
\midrule
DeepSeek-VL-7B-Chat~\cite{lu2024deepseekvl}
& 80.81\% & \textbf{92.77\%} & 68.50\% & 69.60\% & \textbf{88.47\%} & 72.21\% & 84.01\% \\
InstructBLIP-Vicuna-7B~\cite{dai2023instructblip}
& 70.73\% & 88.01\% & 48.88\% & 57.38\% & 85.14\% & 58.15\% & 72.07\% \\
LLaVA-v1.5-7B~\cite{liu2023llava}
& 72.05\% & 85.91\% & 55.59\% & 61.12\% & 82.31\% & 62.26\% & 74.77\% \\
mPLUG-Owl2-7B~\cite{10657415}
& 74.30\% & 88.01\% & 58.35\% & 60.90\% & 86.31\% & 65.69\% & 76.35\% \\
Qwen3-VL-8B-Instruct~\cite{qwen3technicalreport}
& \textbf{82.45\%} & 78.62\% & \textbf{72.72\%} & \textbf{77.86\%} & 84.75\% & \textbf{80.45\%} & \textbf{85.02\%} \\
Qwen2.5-VL-7B-Instruct~\cite{bai2025qwen25vltechnicalreport}
& 79.97\% & 78.55\% & 69.79\% & 72.91\% & 82.80\% & 76.67\% & 84.91\% \\
\bottomrule
\end{tabular}%
\end{table*}

\subsection{Cross-Version Consistency of Model Performance}
To examine whether model capabilities remain consistent across benchmark versions, we conduct a correlation-based analysis comparing model performance on the original benchmark and the dynamically updated benchmark, using Pearson correlation for linear agreement~\cite{pearson1895} and Spearman rank correlation for relative ordering consistency~\cite{spearman1904}.

At the Task-Model Level, each model–capability pair is treated as an independent data point, enabling a fine-grained comparison of capability-level performance across benchmark versions. As shown in Table~\ref{tab:cross_version_correlation}, we observe a stable positive correlation, indicating that models performing well on the original benchmark tend to maintain higher performance on the updated benchmark despite variability across individual capabilities.

At the Model-Averaged Level, performance is aggregated across all capability dimensions for each model before computing correlations. The resulting agreement is notably stronger, suggesting that the relative ordering of models at the overall capability level is largely preserved across benchmark versions. Together, these results indicate that the dynamic update maintains cross-version consistency.

\subsection{Data Contamination Analysis}

To assess potential contamination-related memorization effects in multimodal evaluation benchmarks, we adopt the PaCoST framework~\cite{zhang-etal-2024-pacost}. PaCoST compares model confidence between original questions and their meaning-preserving paraphrases, based on the assumption that memorized or leaked evaluation samples may induce higher confidence on the original formulation than on paraphrased variants.

It is important to note that our analysis should be interpreted as a proxy-based contamination-related signal rather than a direct diagnosis of multimodal data leakage. Since PaCoST measures confidence shifts between original and paraphrased questions, it is primarily sensitive to text-side memorization effects and cannot explicitly distinguish visual leakage, metadata overlap, or exposure through web content. Therefore, our goal is not to determine whether a benchmark is absolutely contaminated, but to compare the relative strength of memorization-related confidence shifts under the same evaluation protocol.

In our experiments, we use Qwen2.5-VL-7B-Instruct as the evaluated model and measure its confidence on both original and paraphrased questions. After producing a multiple-choice answer, the model provides a binary self-assessment of correctness. Based on the decoding probabilities during this decision, we define a confidence score $c \in [0,1]$ that reflects the model's confidence in its predicted answer. For each sample $i$, we compute the confidence difference
\begin{equation}
\Delta_i = c_i^{\text{orig}} - c_i^{\text{para}},
\label{eq:pacost_delta}
\end{equation}
and conduct a one-sided paired $t$-test to evaluate whether $\mathbb{E}[\Delta] > 0$. We further report Cohen’s $d$ as a standardized effect size to quantify the magnitude of confidence shift.

As shown in Table~\ref{tab:contamination_comparison}, both benchmarks exhibit statistically significant confidence shifts under the PaCoST criterion. However, MMBench-Live shows a smaller mean confidence difference and a lower effect size than MMBench. This indicates that, under the same PaCoST-based protocol, MMBench-Live exhibits a weaker memorization-related confidence bias than the original MMBench. We provide additional contamination-related analysis and comparisons with edited benchmark variants in Appendix~\ref{app:additional_contamination}.

\subsection{Task-Level Distribution Consistency}
To further examine whether MMBench-Live preserves the task semantics of the original benchmark, we conduct a complementary task-level distribution analysis. For fairness and comparability, the sample ratio across tasks is explicitly balanced during construction. In addition, since MMBench is entirely composed of multiple-choice questions, MMBench-Live also uses multiple-choice QA pairs with the same number of answer options, ensuring that the answer format is controlled by design.

We analyze the spatial-relation subtask as a representative case, where task-related relation labels can be reliably recovered from QA pairs using an LLM. We compare both the label vocabulary and the frequency distribution between MMBench and MMBench-Live. Based on the recovered labels, MMBench contains 18 label types, with the top-5 labels being north, east, west, northwest, and south. In comparison, MMBench-Live contains 36 label types, with the top-5 labels being north, northeast, east, south, and west. The strong overlap among high-frequency labels indicates that the updated benchmark preserves the core semantic structure of this subtask while expanding label coverage.

We attribute this consistency to the task-related visual patterns and our distribution-consistent update strategy, which help the acquisition agent capture recurring structures in the original benchmark, such as the prevalence of map-based instances in spatial-relation questions.

\subsection{Ablation Study and Analysis.}
\textbf{Effect of Task-Related Visual Patterns and Semantic Feedback.}
We analyze the role of task-related visual patterns and semantic feedback in preserving distributional alignment during dynamic dataset updates. We compare a pattern-aware update pipeline with an unconstrained variant under a matched-sample protocol, evaluating both against the original dataset. Distributional alignment is measured using FID computed on CLIP-ViT-L/14-336 image embeddings and KL divergence computed in a 256-dimensional PCA subspace, with FID averaged over 10 random trials.

As shown in Table~\ref{tab:dist_alignment}, the pattern-aware update achieves lower FID and reduced variance compared to the unconstrained variant, indicating improved global semantic alignment. Beyond global alignment, PCA-based KL divergence reveals structural differences: with task-related visual patterns and semantic feedback enabled, $\mathrm{KL}(A\parallel B)$ slightly exceeds $\mathrm{KL}(B\parallel A)$, yielding a small positive $\Delta\mathrm{KL}$ and suggesting that the updated data form a compact sub-distribution supported by the original dataset. In contrast, removing these mechanisms reverses this relation, resulting in a negative $\Delta\mathrm{KL}$ that reflects reduced coverage and increased structural drift. Overall, these results highlight the importance of task-related visual patterns and semantic feedback in preserving distributional alignment during dynamic benchmark updates.

\textbf{Does the Effectiveness of the Pipeline Depend on GPT?}
To assess whether the proposed pipeline depends on GPT-specific capabilities, we replace the large-model components in task specification and data acquisition with Qwen3 and Qwen3-VL. As shown in Table~\ref{tab:dynamic_mmbench_l2_qwen}, the resulting MMBench-Live still exhibits clear performance separation across model families, indicating that the pipeline remains effective without relying on GPT models.

We further examine cross-version consistency between the original benchmark and the Qwen-based dynamic benchmark. Stable positive correlations are observed at both the Task-Model level (Pearson $r = 0.4574$, Spearman $\rho = 0.4408$) and the Model-Averaged level (Pearson $r = 0.9084$, Spearman $\rho = 0.8286$), suggesting that overall model ordering remains largely preserved. These results demonstrate that the proposed method can stably construct discriminative and cross-version-consistent benchmarks even when GPT models are fully replaced.

\section{Limitations and Discussion}
Despite its effectiveness in mitigating data contamination and improving evaluation timeliness, the proposed dynamic benchmark construction framework has limitations. Dynamic updates cannot fully eliminate implicit memorization of high-frequency visual concepts, particularly for recurring objects, scenes, or styles. Moreover, the quality of automatically constructed evaluation instances is bounded by the capabilities of the underlying agents and foundation models, and subtle noise may persist under visual ambiguity or underspecified constraints. From an extensibility perspective, the current task-grounded paradigm prioritizes cross-version comparability and stable evaluation objectives, which facilitates tracking progress along predefined capabilities but limits task-space expansion. Future work will explore complementary task-expansive update strategies that introduce new task formulations alongside task-grounded updates to enable more adaptive evaluation of evolving VLMs.

\section{Conclusion}
This work revisits multimodal benchmark construction under the challenges of temporal obsolescence, data contamination, and high maintenance cost in static evaluation paradigms. We propose a task-centered, multi-agent framework that treats benchmark evolution as a task-guided dataset construction process, integrating structured benchmark descriptions, task-grounded data acquisition with feedback control, and automated instance construction with executable verification. Instantiated on MMBench, the resulting MMBench-Live enables fully automated updates with temporally recent and semantically aligned evaluation instances while preserving cross-version comparability. Experiments demonstrate high QA correctness, improved robustness to contamination, and stable performance correlations across versions. The proposed framework is general and extensible to other multimodal benchmarks, providing a practical paradigm for sustainable evaluation of evolving vision–language models.

\section*{Impact Statement}

This paper presents work whose goal is to advance the field of Machine
Learning. There are many potential societal consequences of our work, none
which we feel must be specifically highlighted here.


\bibliography{example_paper}
\bibliographystyle{icml2026}

%%%%%%%%%%%%%%%%%%%%%%%%%%%%%%%%%%%%%%%%%%%%%%%%%%%%%%%%%%%%%%%%%%%%%%%%%%%%%%%
%%%%%%%%%%%%%%%%%%%%%%%%%%%%%%%%%%%%%%%%%%%%%%%%%%%%%%%%%%%%%%%%%%%%%%%%%%%%%%%
% APPENDIX
%%%%%%%%%%%%%%%%%%%%%%%%%%%%%%%%%%%%%%%%%%%%%%%%%%%%%%%%%%%%%%%%%%%%%%%%%%%%%%%
%%%%%%%%%%%%%%%%%%%%%%%%%%%%%%%%%%%%%%%%%%%%%%%%%%%%%%%%%%%%%%%%%%%%%%%%%%%%%%%
\newpage
\appendix
\onecolumn

% =========================
% A. Representative QA Examples
% =========================
\section{Failure Analysis}
\label{app:failure_analysis}
Although executable reasoning verification substantially improves the reliability of automatically generated QA pairs, it does not eliminate all errors. We observe that the remaining failures are often not caused by low-level perception, but by the final reasoning stage that aggregates tool-derived evidence.

\begin{figure*}[h]
\centering
\includegraphics[width=0.95\linewidth]{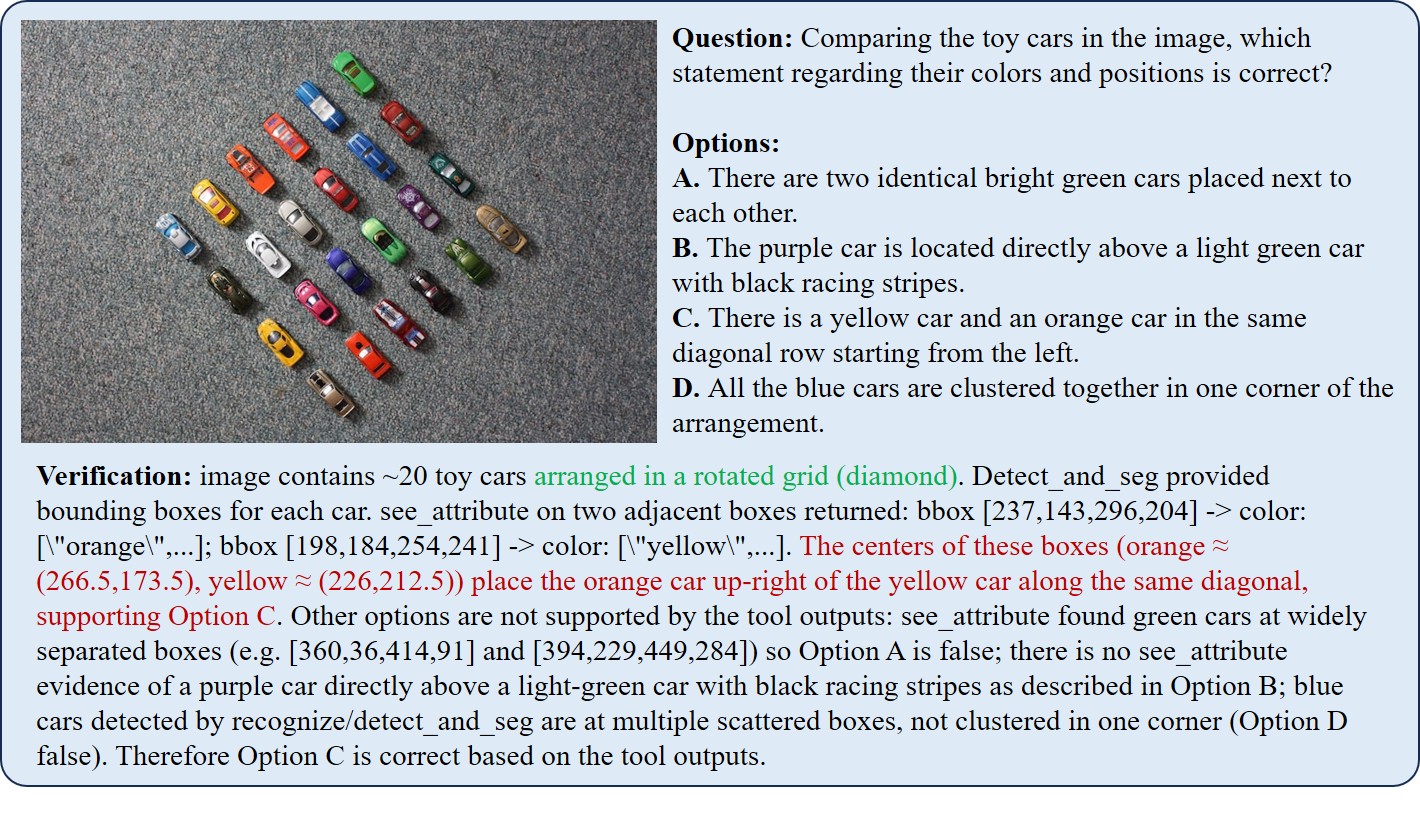}
\caption{
A representative failure case of executable reasoning verification.
}
\label{fig:failure_case}
\end{figure*}

Figure~\ref{fig:failure_case} shows a representative failure case from a spatial-relation question about toy cars. The question asks: “Comparing the toy cars in the image, which statement regarding their colors and positions is correct?” One candidate option states that “There is a yellow car and an orange car in the same diagonal row starting from the left.” In this case, the perception modules correctly identify the toy cars and the rotated grid layout. The upstream tools also return the relevant color attributes and bounding boxes. However, the final reasoning stage incorrectly concludes that the yellow and orange cars lie on the same diagonal.

A closer inspection shows that the error comes from interpreting the rotated grid structure. Under the rotated reference frame, the upper-right direction should be treated as a row direction rather than a diagonal direction. Therefore, the two cars are not on the same diagonal, even though their relative positions can appear diagonal under the original image coordinate system. This indicates that the main error source lies in relation reasoning over structured tool outputs, rather than in object detection, attribute recognition, or localization.

This case highlights both the benefit and limitation of executable verification. On the one hand, the tool chain makes the failure traceable by separating perception results from final reasoning. On the other hand, the verifier may still make incorrect relational inferences when the scene involves ambiguous reference frames, rotated layouts, or non-standard spatial structures. Future work may incorporate explicit geometric normalization or rule-based spatial consistency checks to further reduce such errors.

\section{Additional Contamination Analysis}
\label{app:additional_contamination}
We provide additional PaCoST-based contamination-related analysis to further compare MMBench-Live with VLB-style~\cite{yang2025dynamic} image editing variants. Here, Adding, Elimination, and Outpainting denote edited variants constructed by adding visual content, removing visual content, and extending image regions, respectively. 

As shown in Table~\ref{tab:additional_contamination}, the edited variants, including adding, elimination, and outpainting, exhibit substantially larger confidence shifts and effect sizes than MMBench-Live. In contrast, MMBench-Live shows a much smaller mean confidence shift and Cohen's $d$. Combined with temporal isolation in data acquisition, these results suggest that our live updating paradigm is more effective than image-editing-based variants in reducing contamination-related memorization signals.

We emphasize that PaCoST should be interpreted as a proxy-based analysis rather than a definitive contamination detector. Although the results provide suggestive evidence, they are insufficient on their own to draw strong conclusions about multimodal contamination, especially given the limited transparency of large-scale multimodal pretraining corpora.

\begin{table}[h]
\centering
\caption{Additional PaCoST-based confidence-shift comparison between edited variants and MMBench-Live.}
\label{tab:additional_contamination}
\begin{tabular}{lcccc}
\toprule
Benchmark Variant & Mean $\Delta$ & Std $\Delta$ & $p$ (one-sided) & Cohen's $d$ \\
\midrule
Adding~\cite{yang2025dynamic} & 0.042 & 0.187 & $6.1\times10^{-14}$ & 0.224 \\
Elimination~\cite{yang2025dynamic} & 0.038 & 0.200 & $3.7\times10^{-12}$ & 0.191 \\
Outpainting~\cite{yang2025dynamic} & 0.042 & 0.205 & $1.1\times10^{-13}$ & 0.205 \\
MMBench-Live & \textbf{0.006} & \textbf{0.102} & $2.9\times10^{-6}$ & \textbf{0.059} \\
\bottomrule
\end{tabular}
\end{table}

\section{Task Abbreviations}
\label{sec:appendix:abbrev}
\begin{table}[H]
\centering
\small
\setlength{\tabcolsep}{10pt}
\begin{tabular}{ll}
\toprule
\textbf{Abbrev.} & \textbf{Task Name} \\
\midrule
IT   & Image Topic \\
IQ   & Image Quality \\
IE   & Image Emotion \\
IS   & Image Scene \\
IST  & Image Style \\
ARc  & Attribute Recognition \\
OL   & Object Localization \\
CR   & Celebrity Recognition \\
OCR  & Optical Character Recognition \\
SR   & Spatial Relationship \\
AC   & Attribute Comparison \\
ActR & Action Recognition \\
IR   & Identity Reasoning \\
FR   & Function Reasoning \\
PPR  & Physical Property Reasoning \\
FP   & Future Prediction \\
SITU & Structuralized Image-Text Understanding \\
SoR  & Social Relation \\
PhR  & Physical Relation \\
NaR  & Nature Relation \\
\bottomrule
\end{tabular}
\caption{Task abbreviations and corresponding full task names.}
\label{tab:task_abbrev_single}
\end{table}

\section{MMBench-Live Representative QA Examples}
\label{sec:appendix:qa}

% ---------- GPT-based ----------
\subsection{QA Examples on the GPT-based Version}
\label{sec:appendix:qa:gpt}

\begin{center}
  \includegraphics[width=\linewidth]{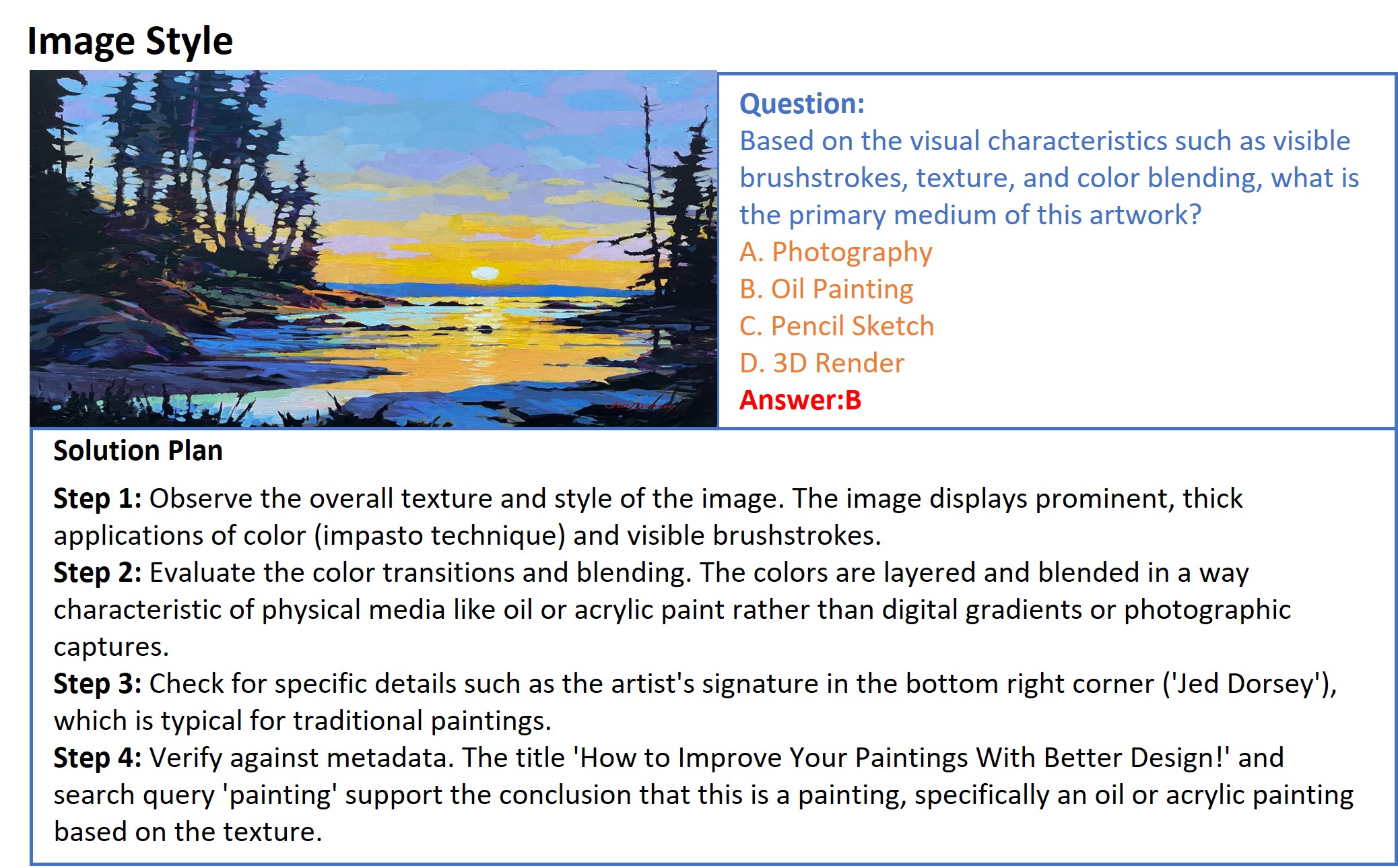}

  \includegraphics[width=\linewidth]{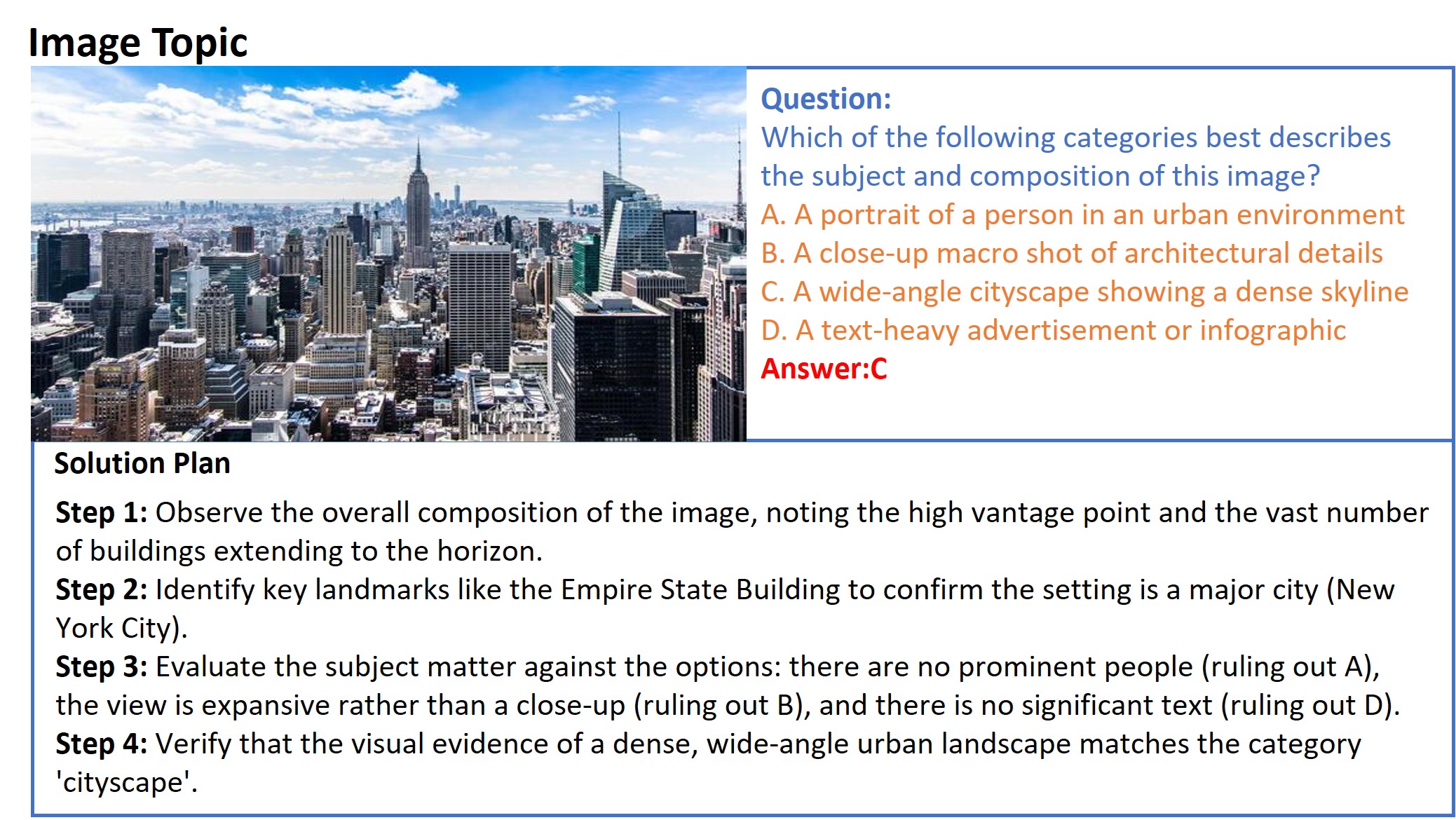}
\end{center}
\begin{center}
  \includegraphics[width=\linewidth]{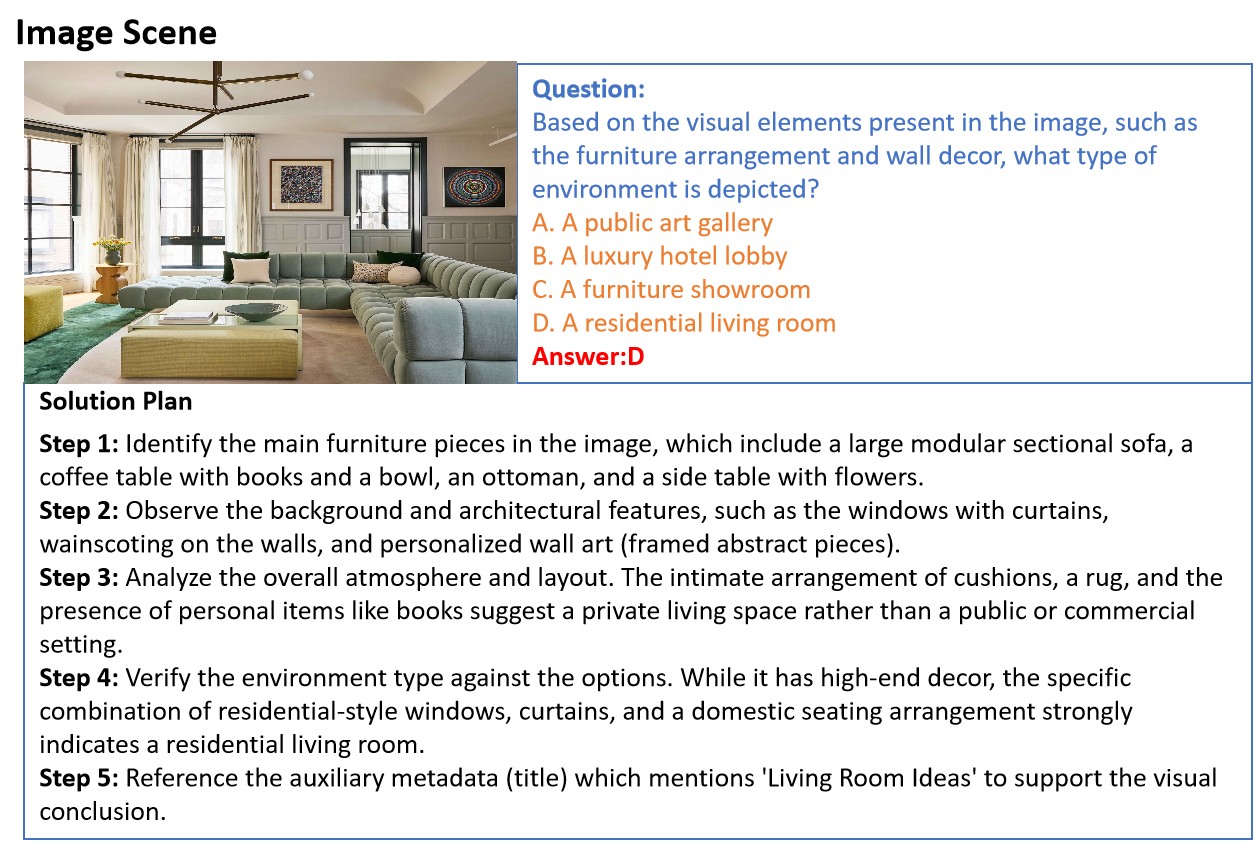}

  \includegraphics[width=\linewidth,height=0.45\textheight]{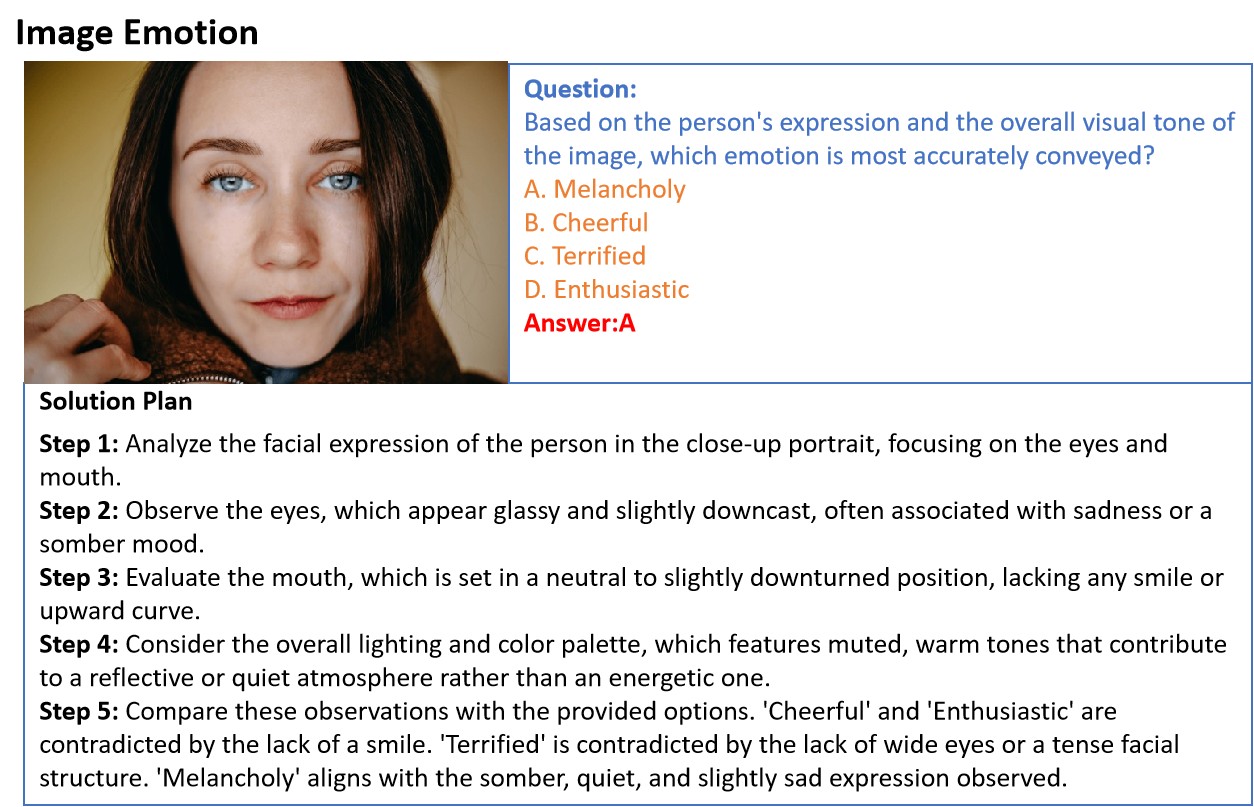}
\end{center}
\begin{center}
  \includegraphics[width=\linewidth]{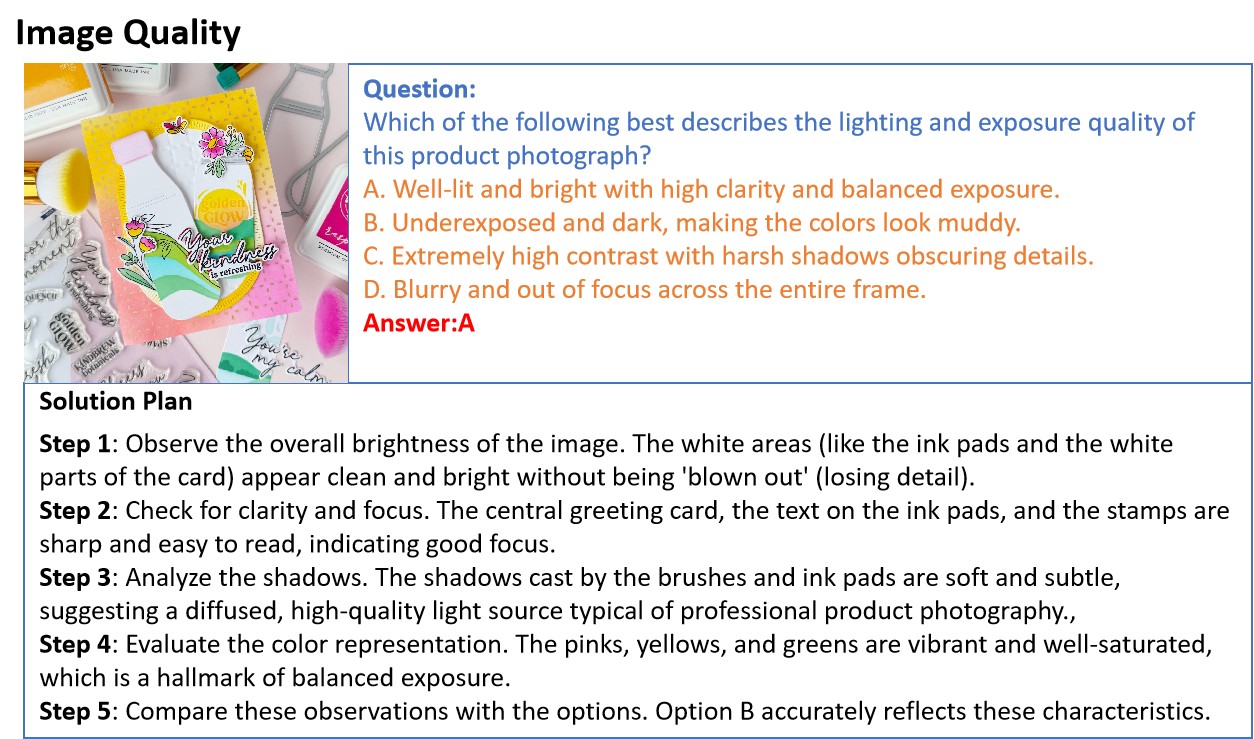}

  \includegraphics[width=\linewidth]{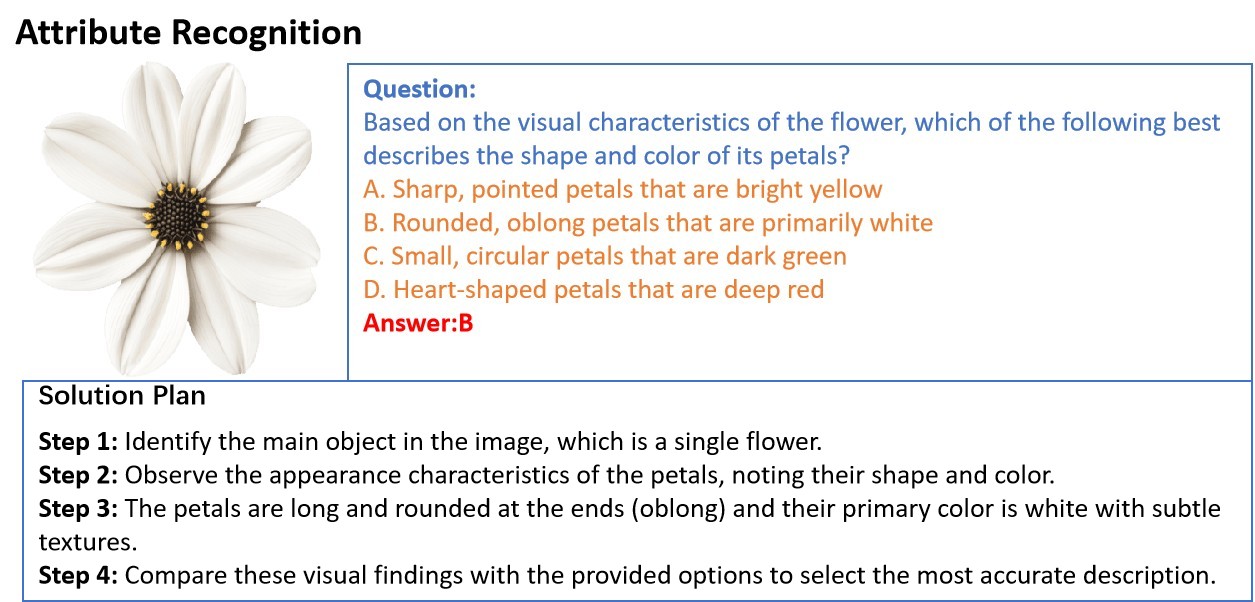}
\end{center}

\begin{center}
  \includegraphics[width=\linewidth]{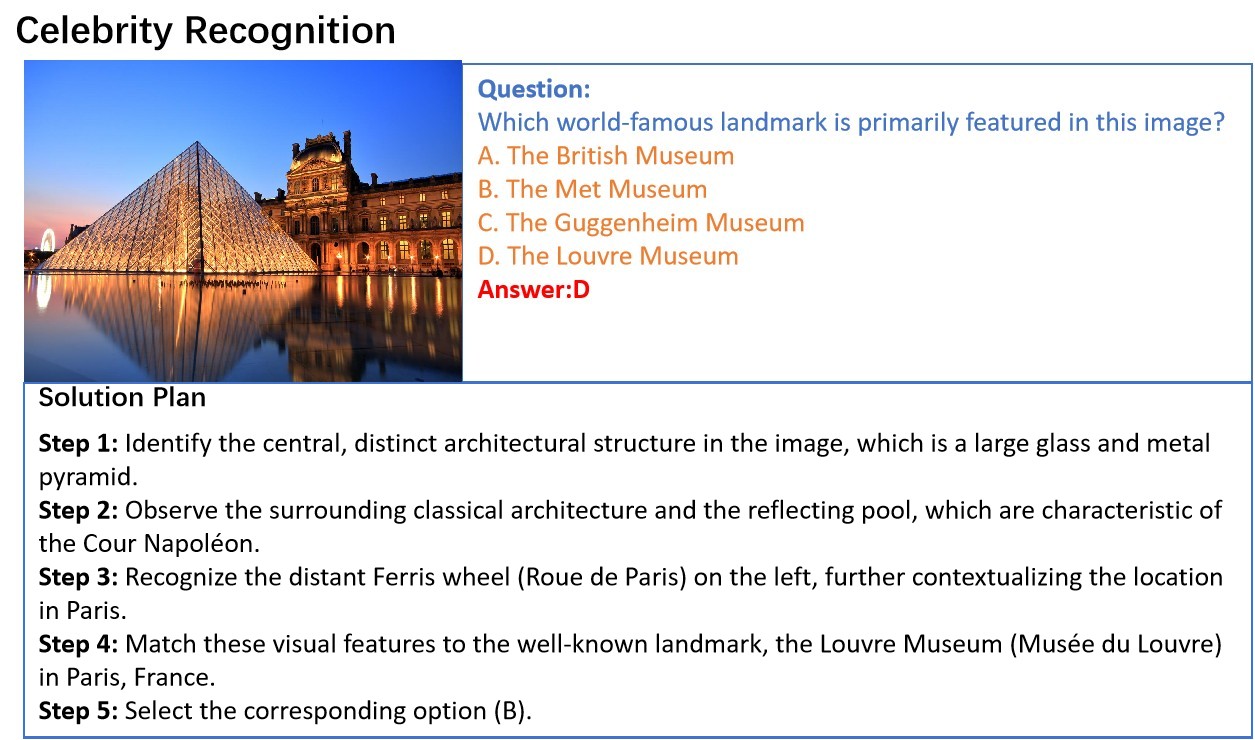}

  \includegraphics[width=\linewidth,height=0.45\textheight]{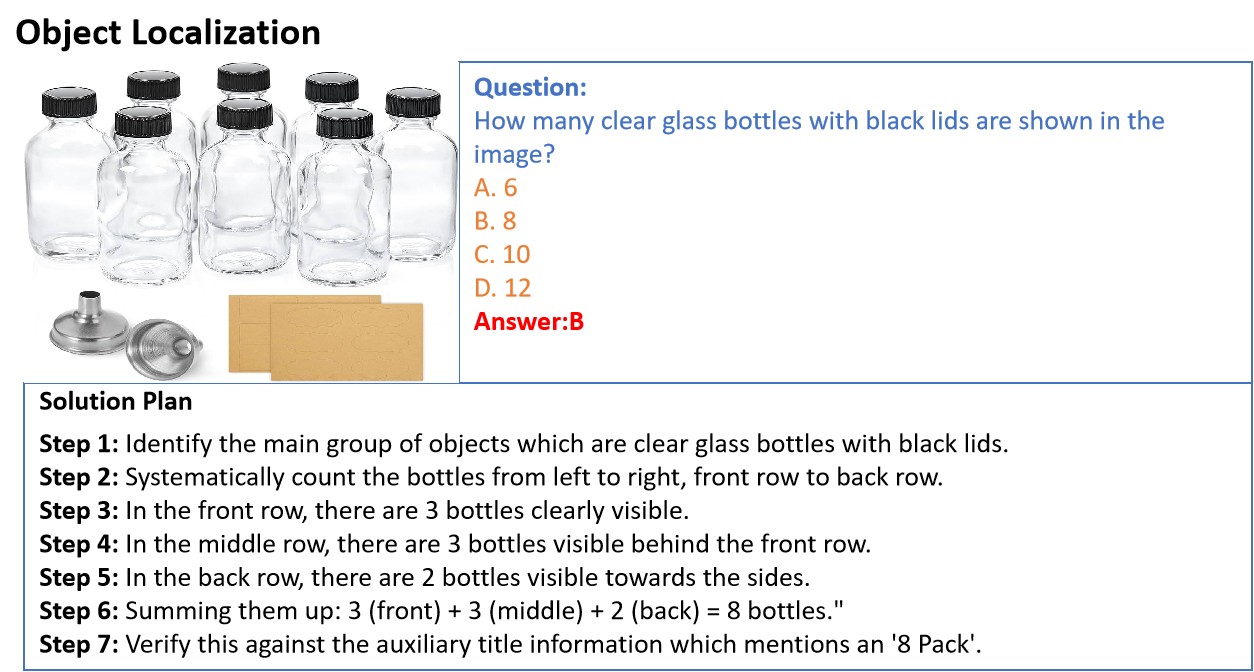}
\end{center}
\begin{center}
  \includegraphics[width=\linewidth]{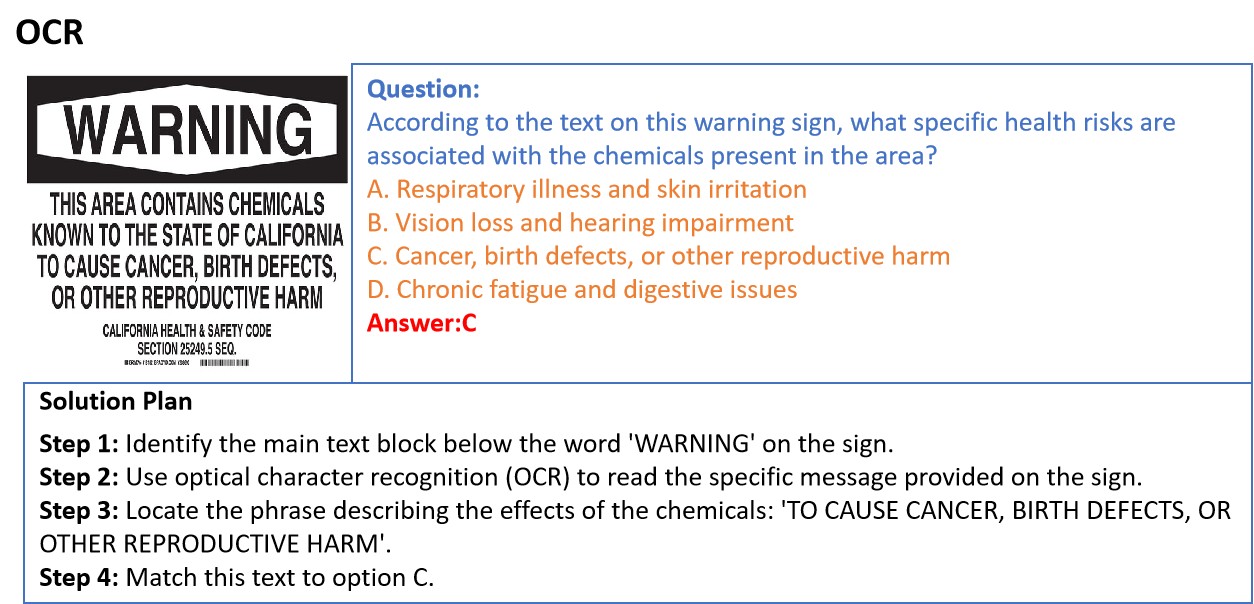}

  \includegraphics[width=\linewidth,height=0.45\textheight]{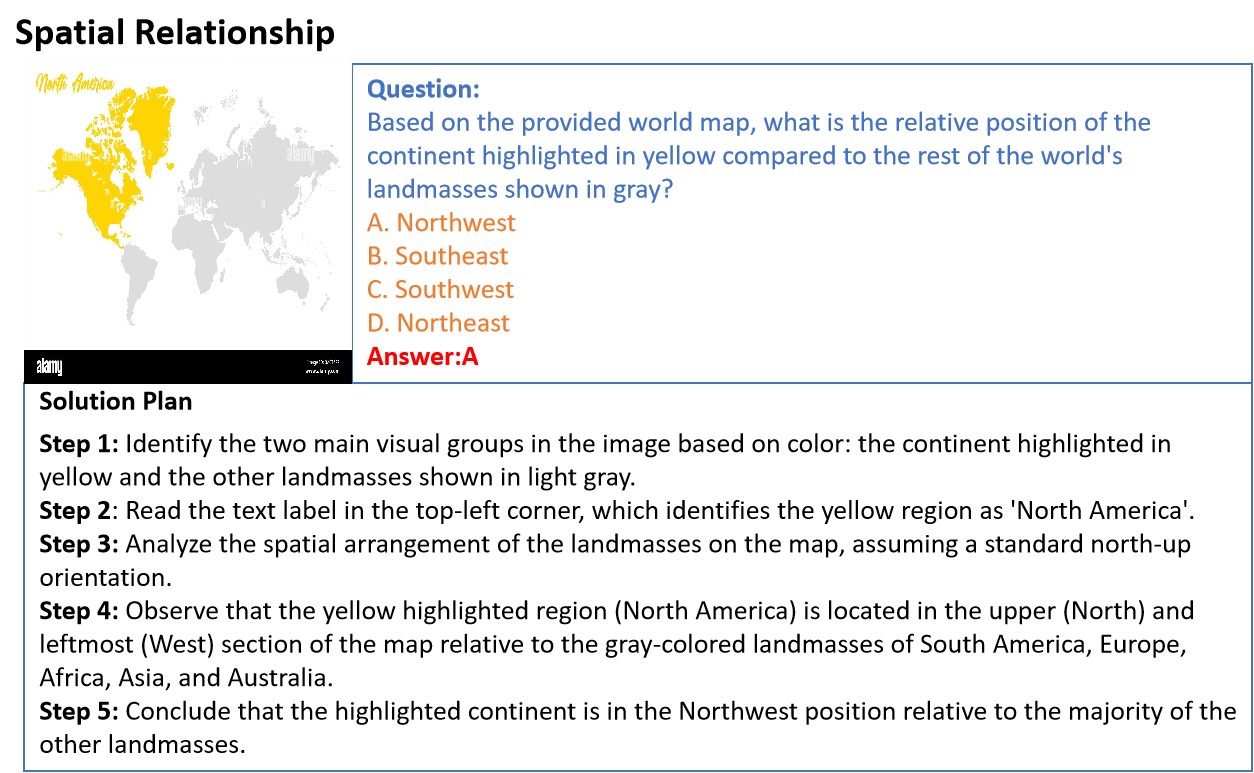}
\end{center}
\begin{center}
  \includegraphics[width=\linewidth]{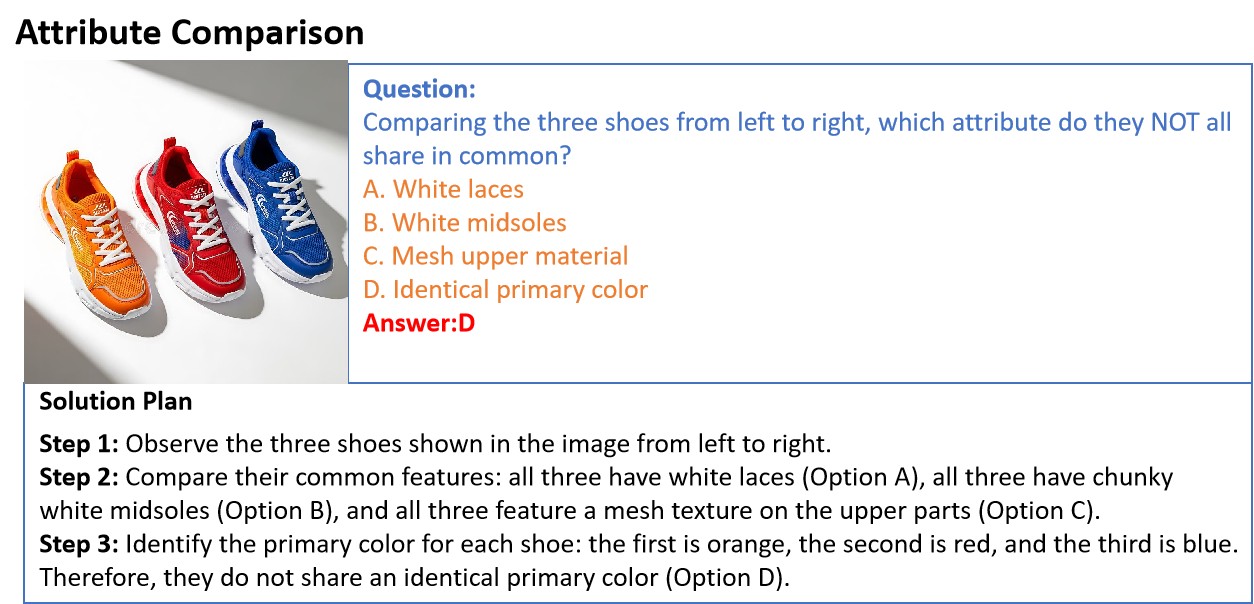}

  \includegraphics[width=\linewidth,height=0.45\textheight]{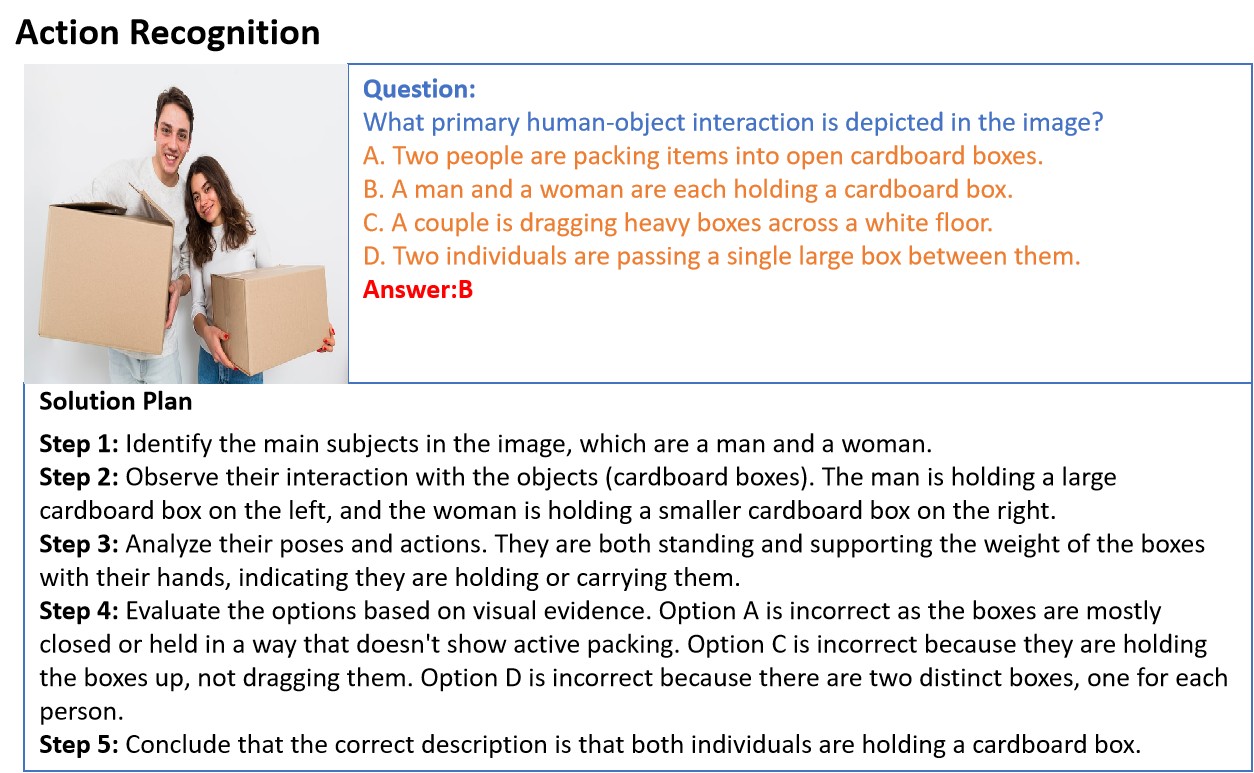}
\end{center}

\begin{center}
  \includegraphics[width=\linewidth,height=0.45\textheight]{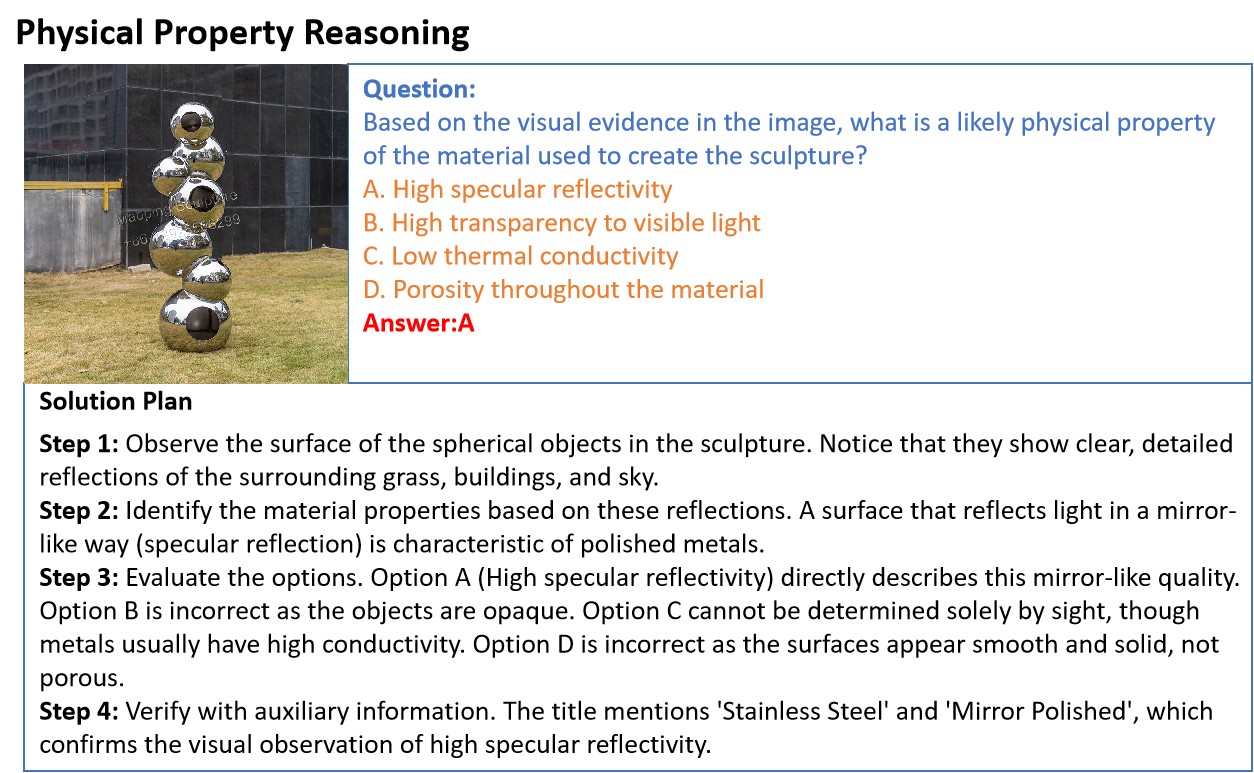}

  \includegraphics[width=\linewidth,height=0.45\textheight]{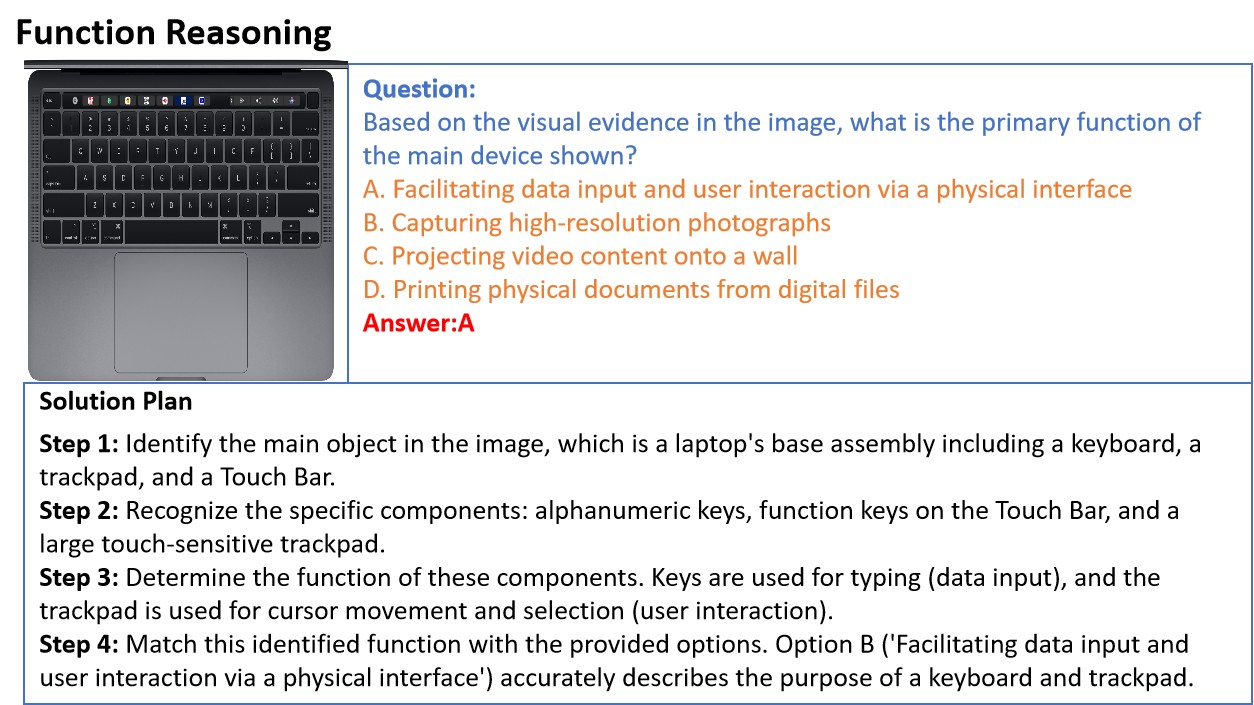}
\end{center}

\begin{center}
  \includegraphics[width=\linewidth,height=0.45\textheight]{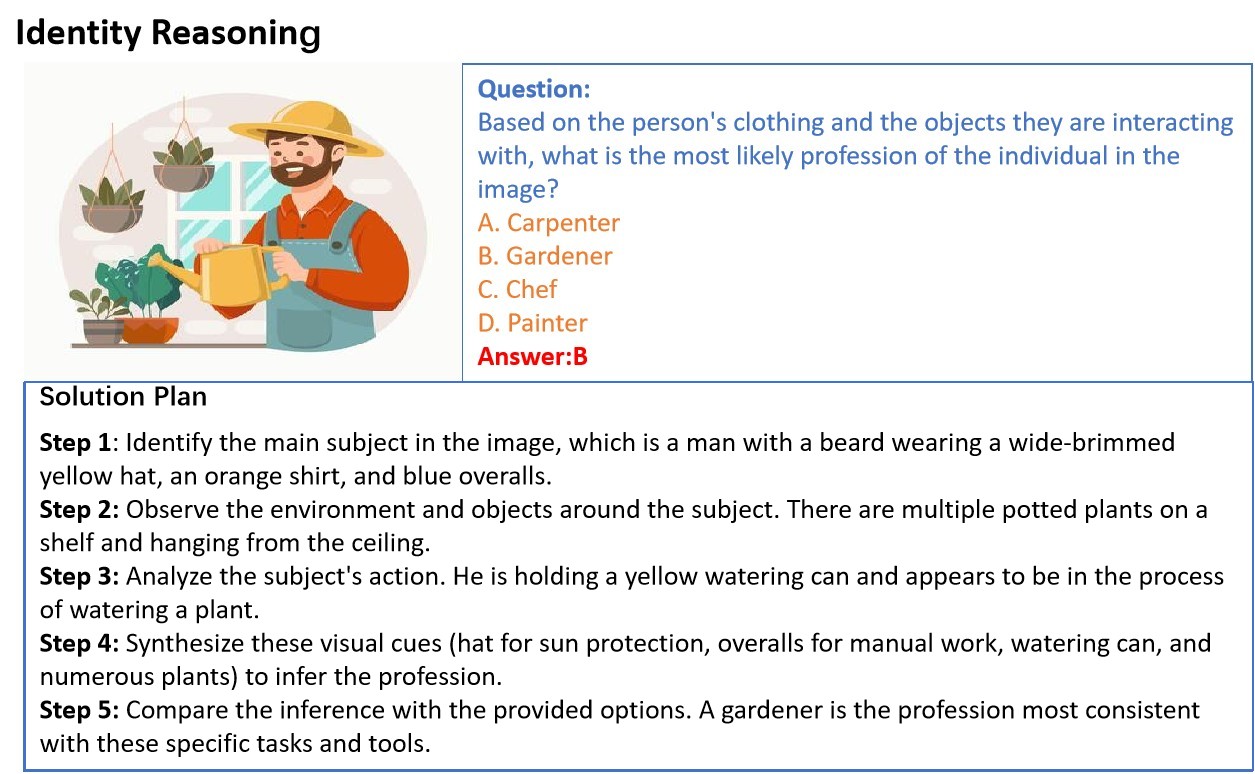}

  \includegraphics[width=\linewidth,height=0.45\textheight]{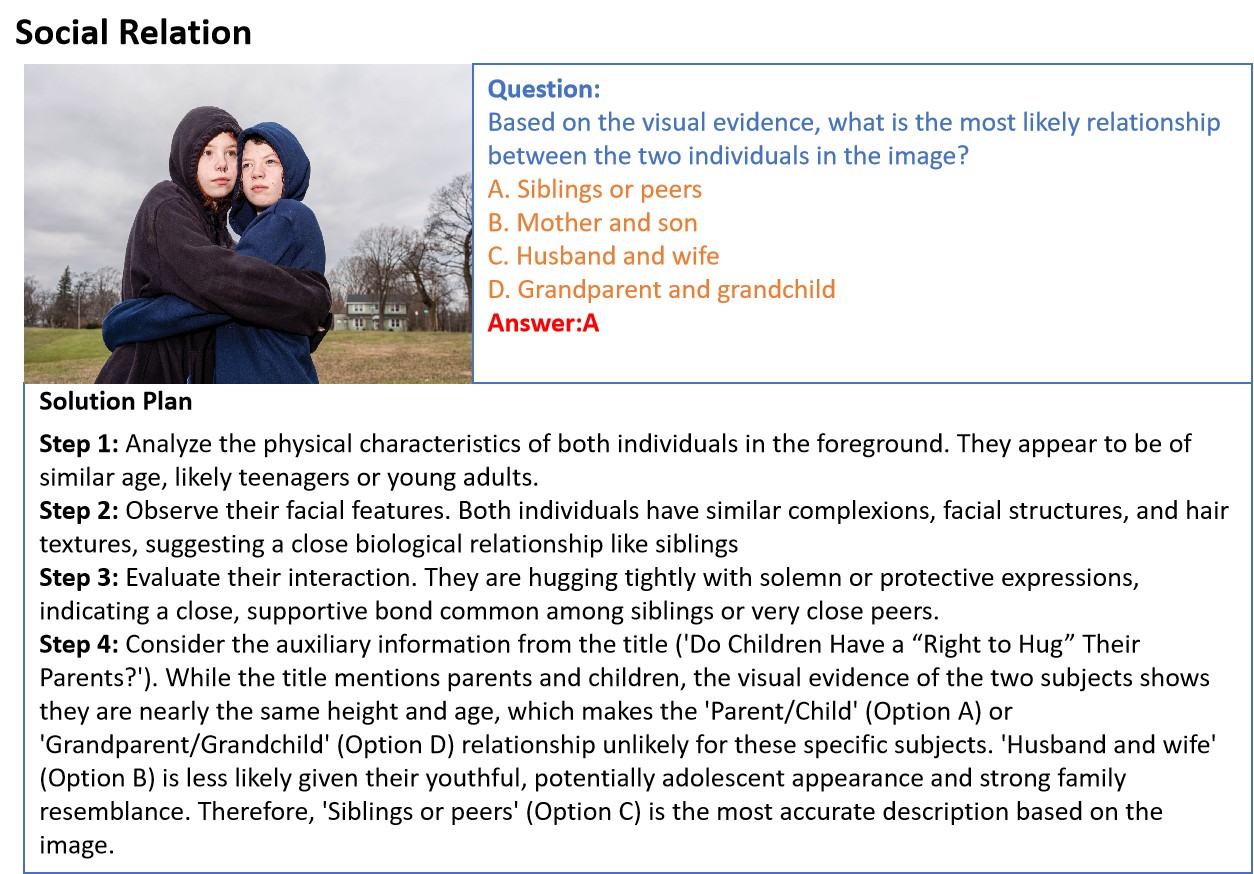}
\end{center}
\begin{center}
  \includegraphics[width=\linewidth,height=0.45\textheight]{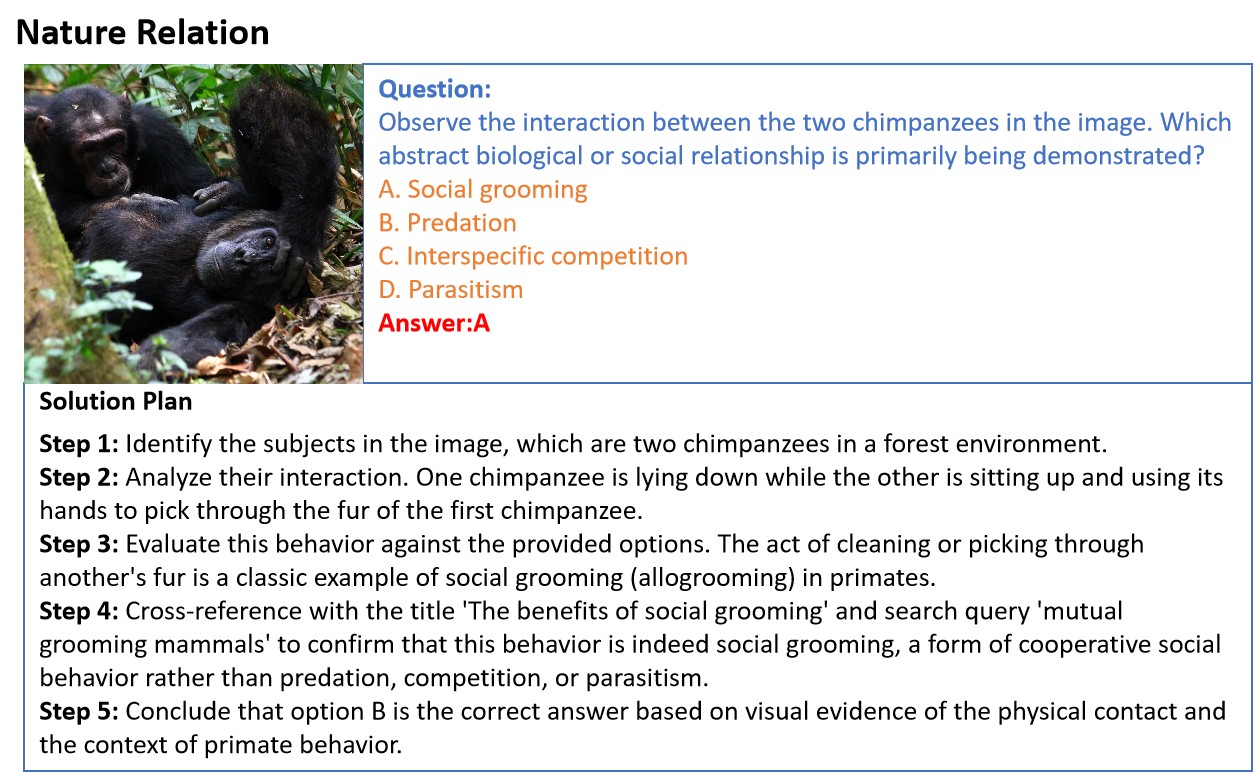}

  \includegraphics[width=\linewidth,height=0.45\textheight]{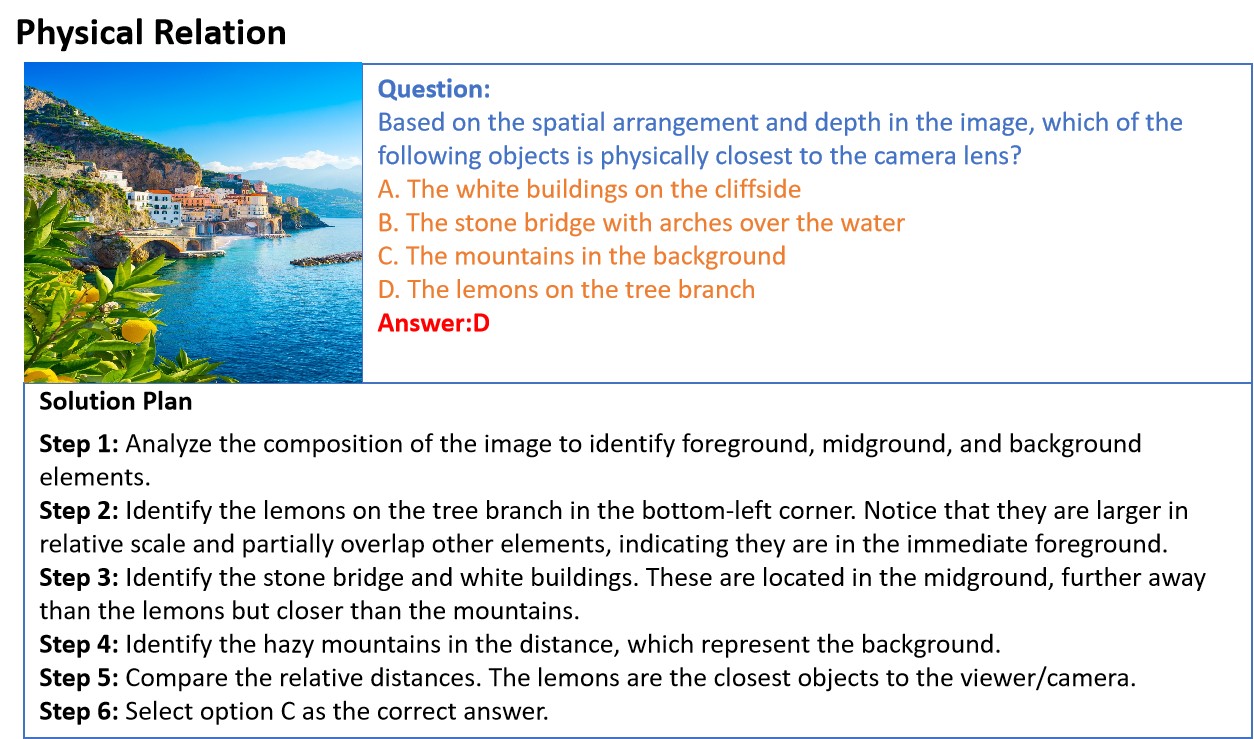}
\end{center}

\begin{center}
  \includegraphics[width=\linewidth,height=0.45\textheight]{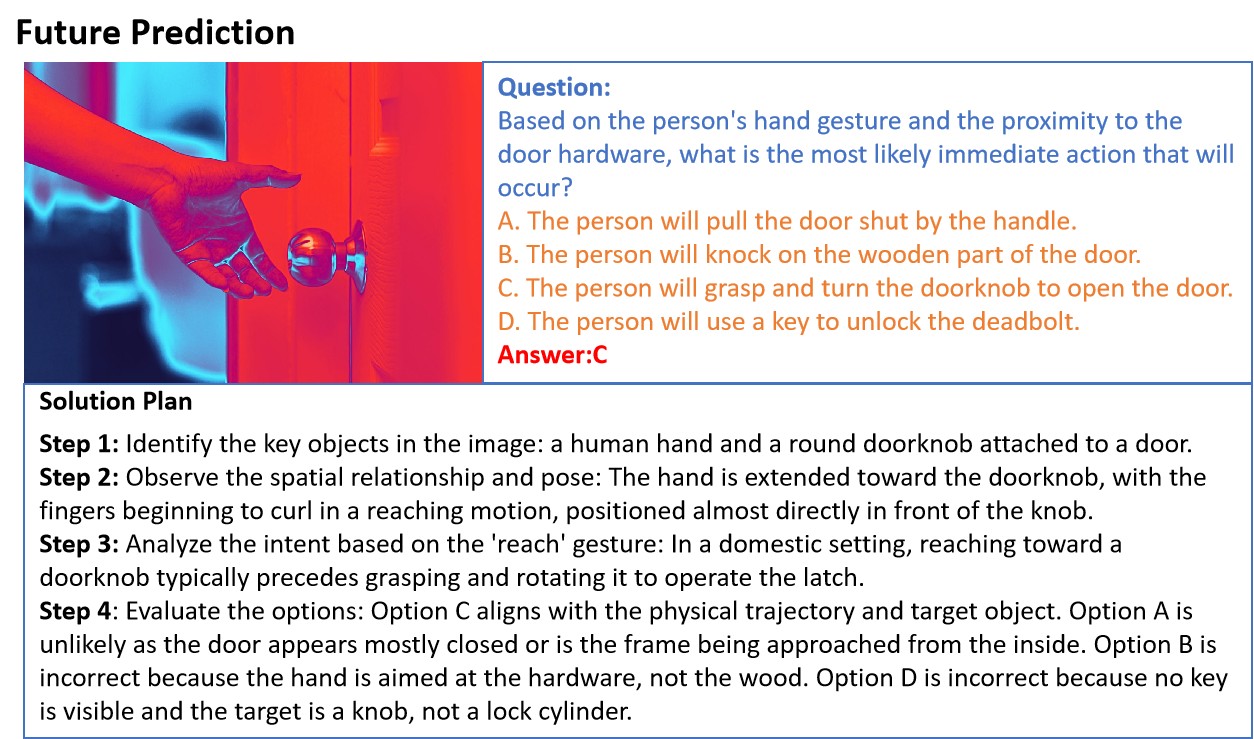}

  \includegraphics[width=\linewidth,height=0.45\textheight]{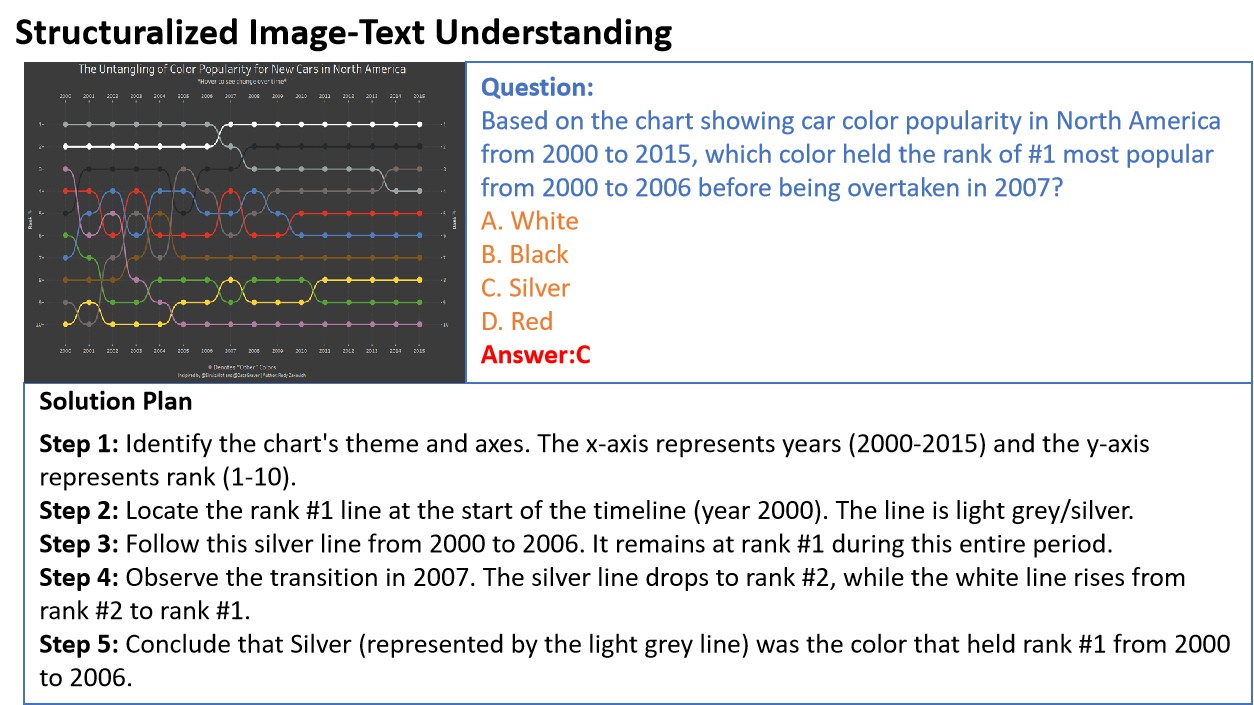}
\end{center}

% ---------- Qwen-based ----------
\subsection{QA Examples on the Qwen-based Version}
\label{sec:appendix:qa:qwen}

\begin{center}
  \includegraphics[width=\linewidth]{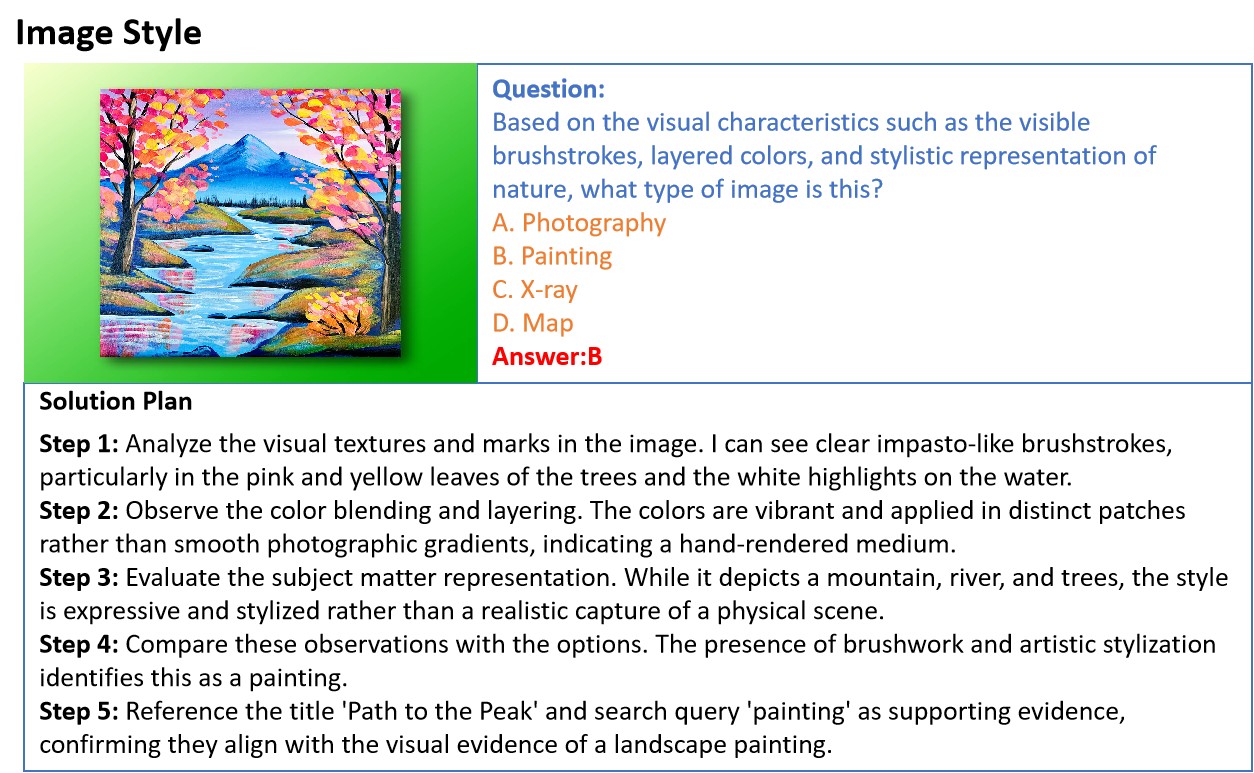}

  \includegraphics[width=\linewidth]{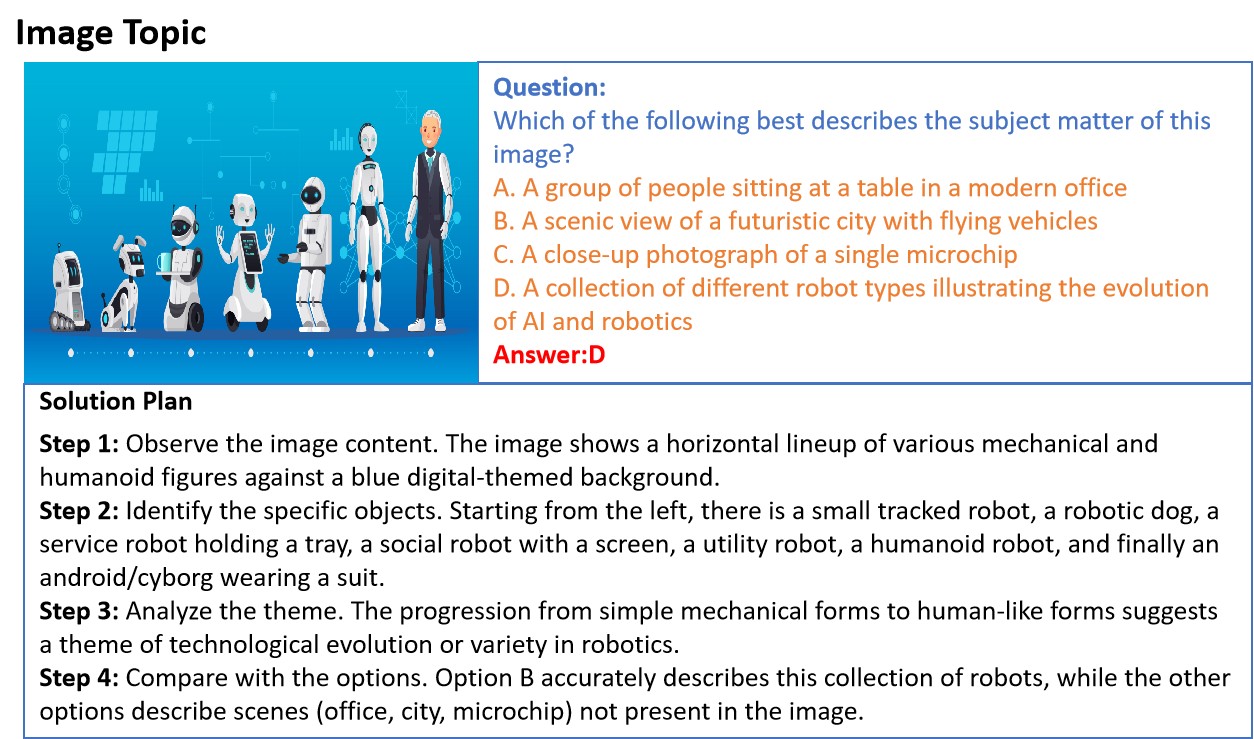}
\end{center}
\begin{center}
  \includegraphics[width=\linewidth]{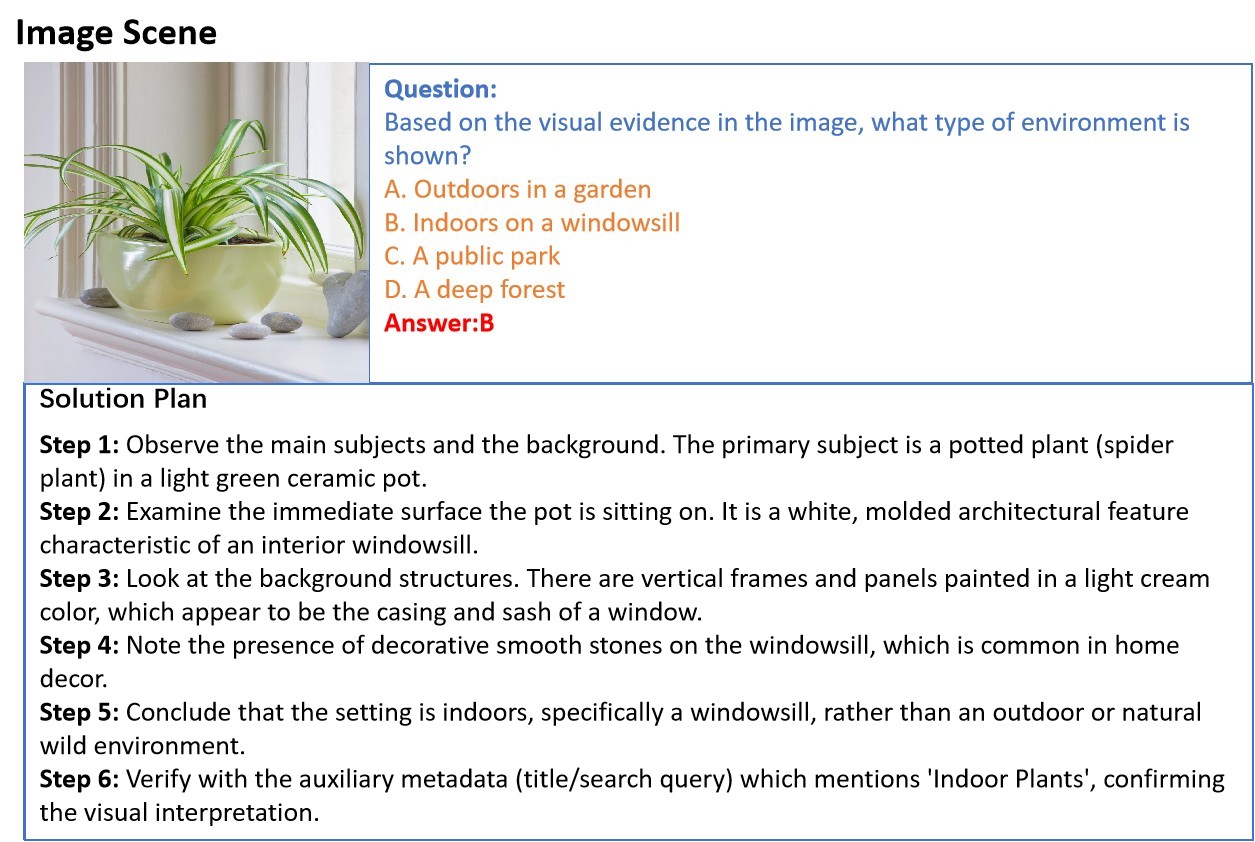}

  \includegraphics[width=\linewidth,height=0.45\textheight]{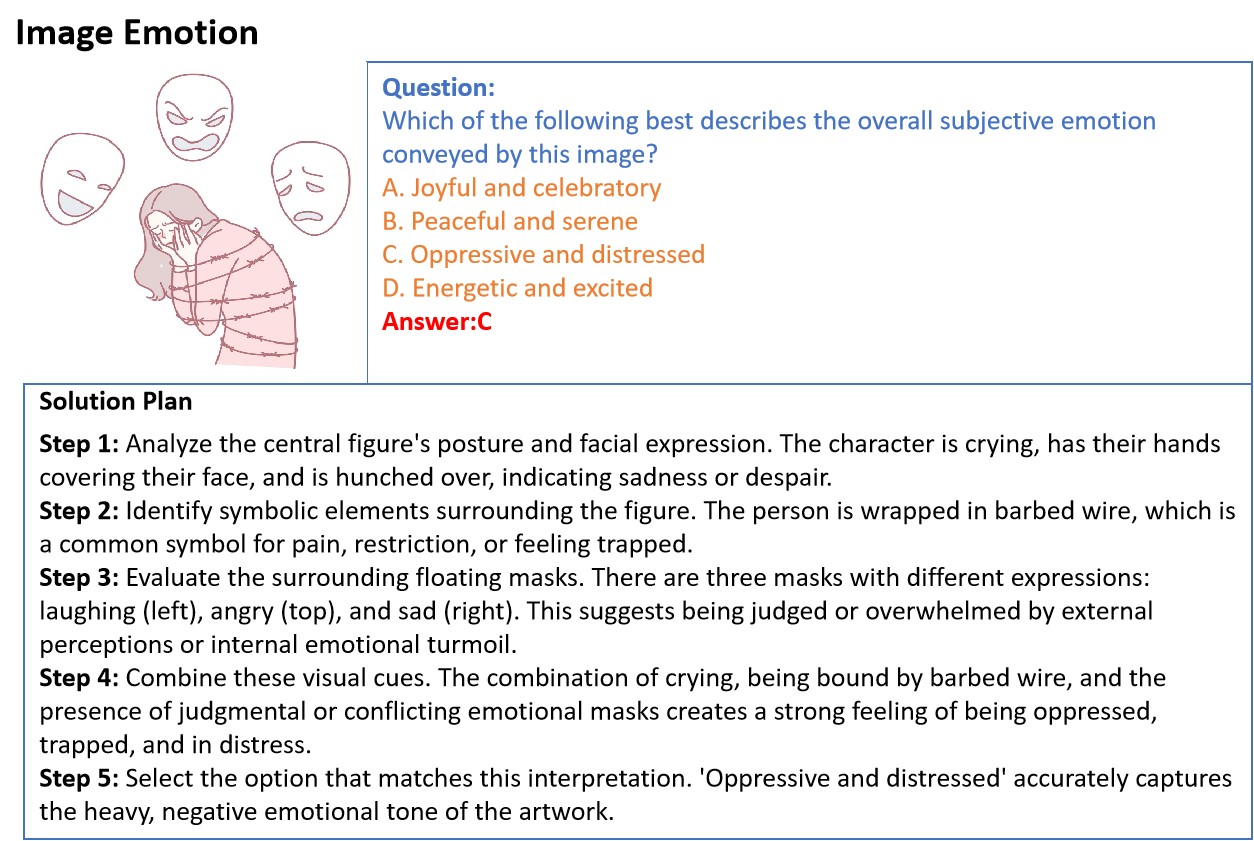}
\end{center}
\begin{center}
  \includegraphics[width=\linewidth]{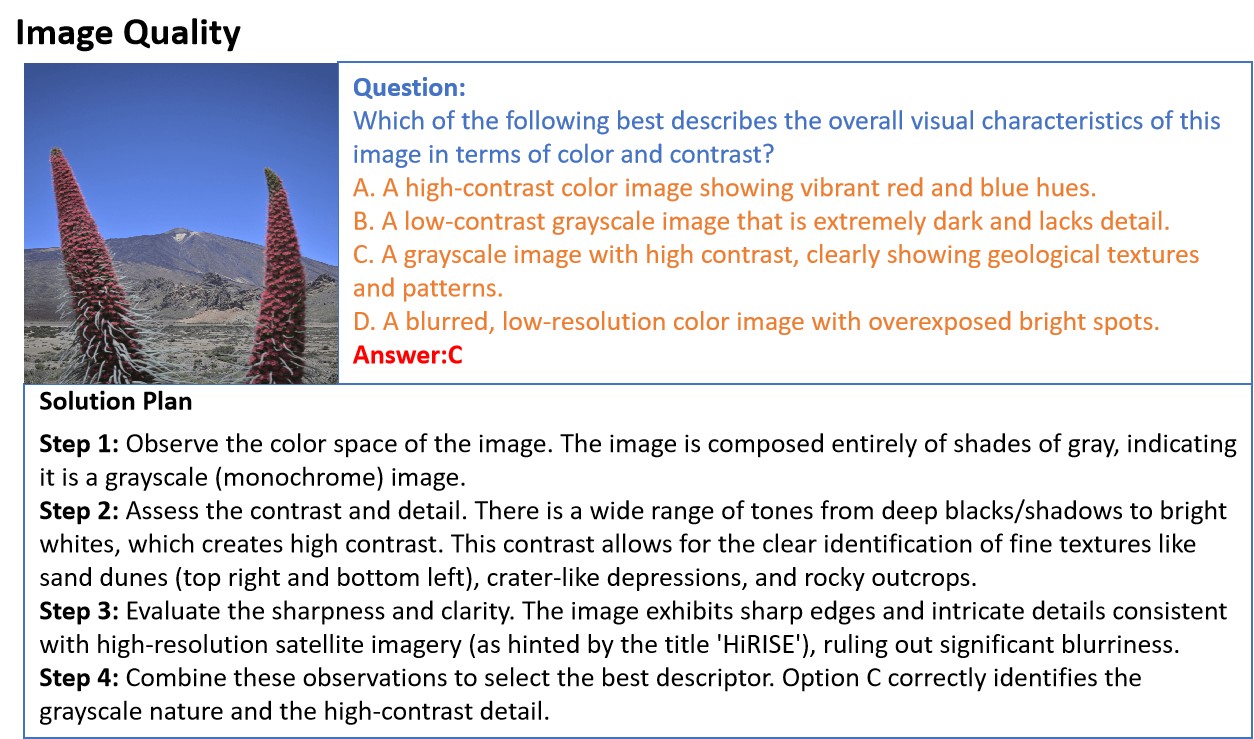}

  \includegraphics[width=\linewidth]{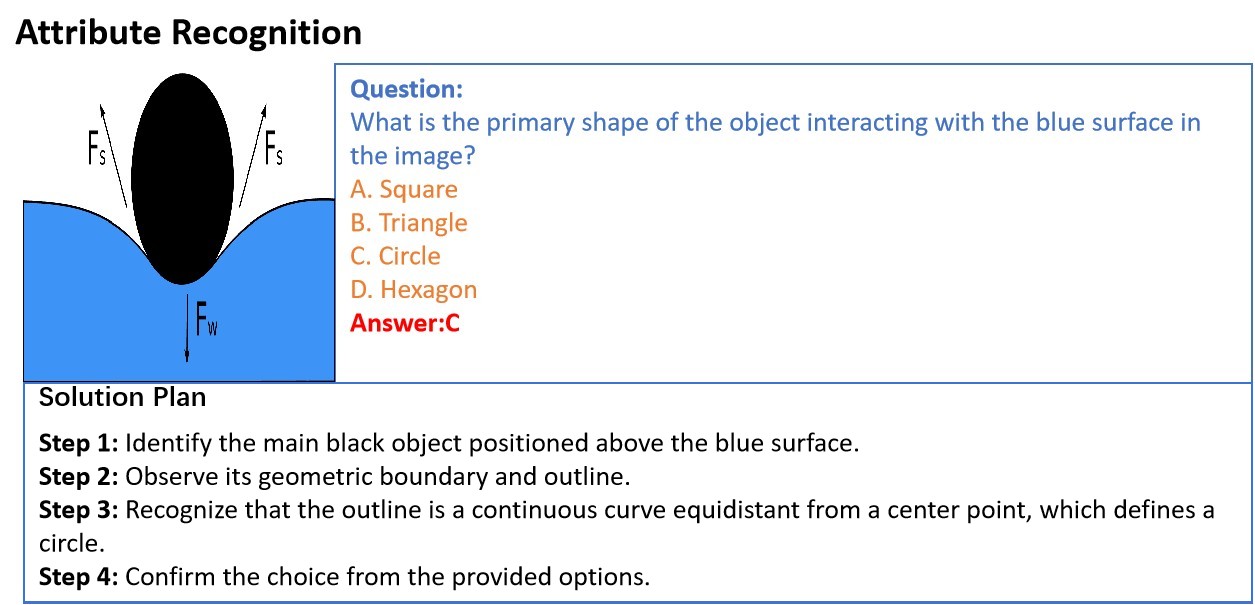}
\end{center}

\begin{center}
  \includegraphics[width=\linewidth]{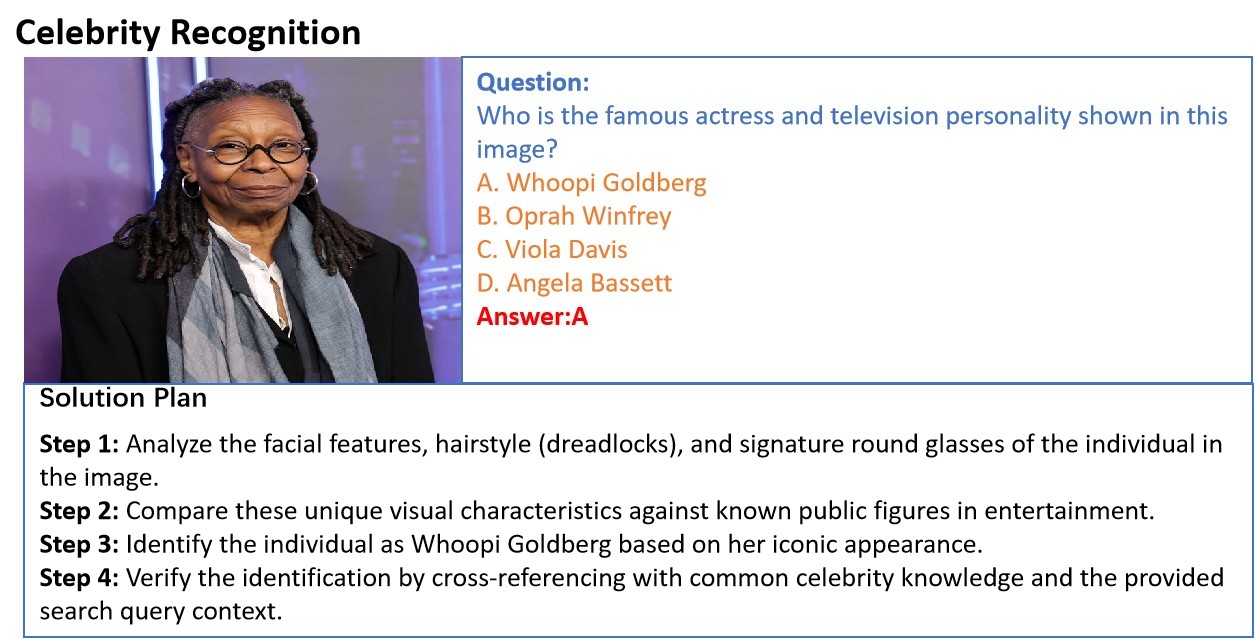}

  \includegraphics[width=\linewidth,height=0.45\textheight]{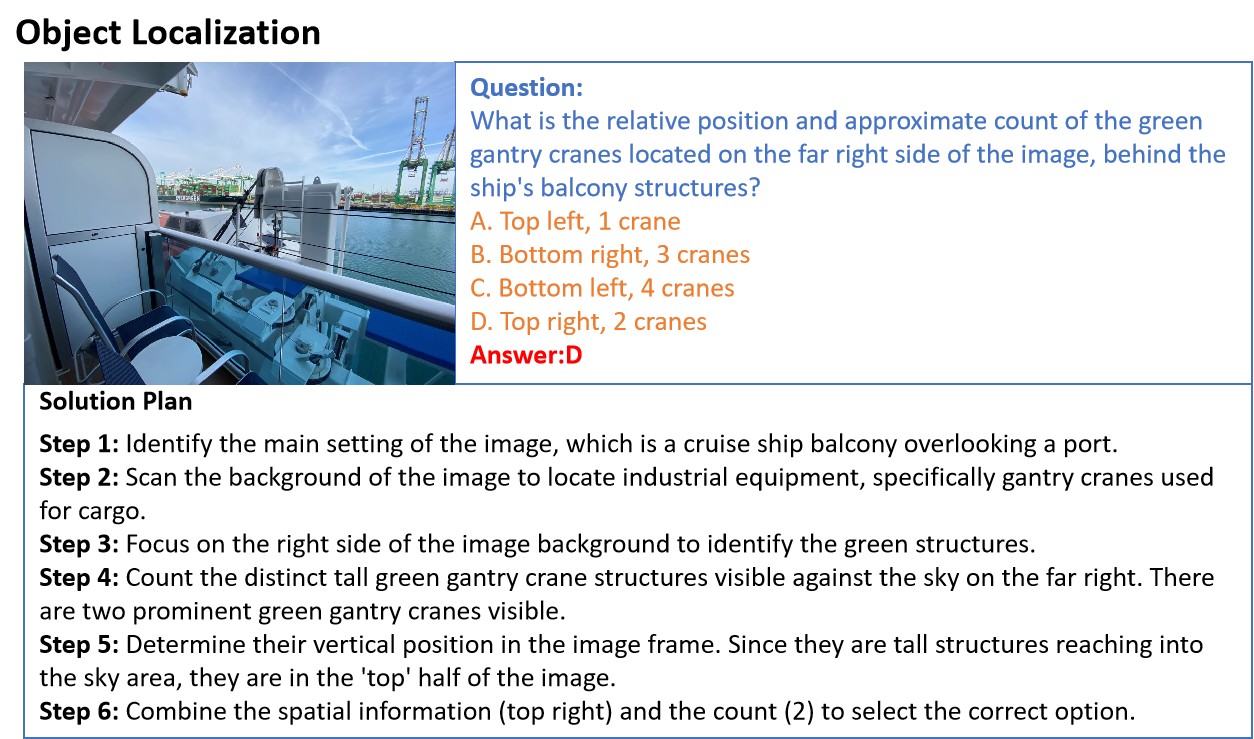}
\end{center}
\begin{center}
  \includegraphics[width=\linewidth]{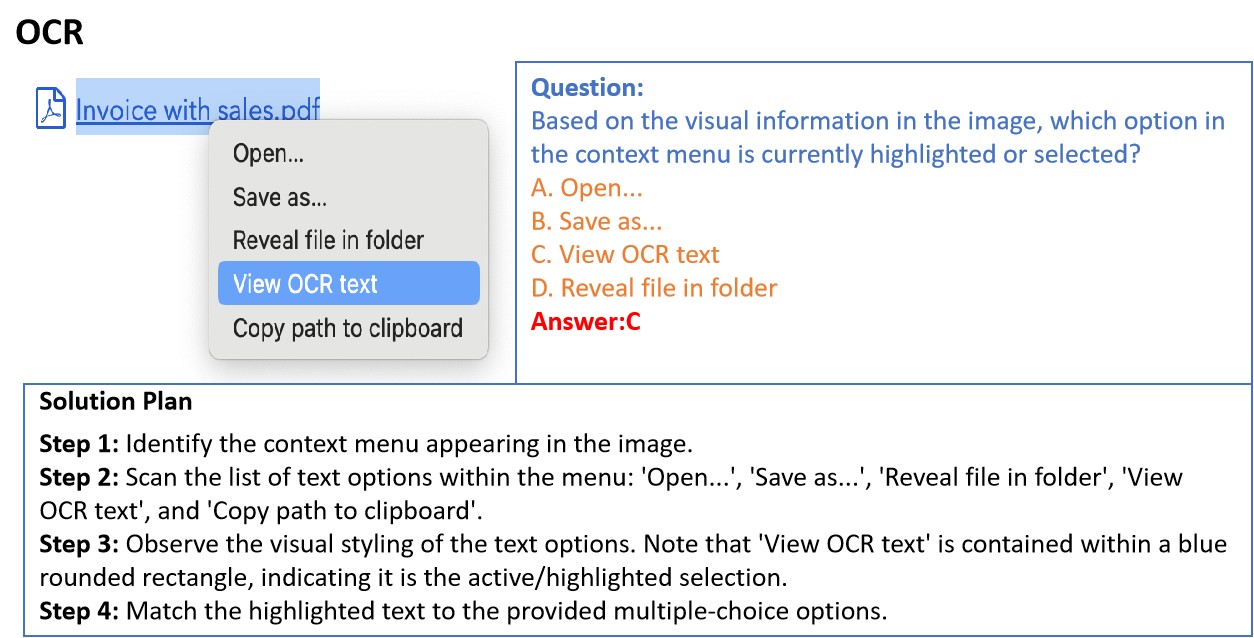}

  \includegraphics[width=\linewidth,height=0.45\textheight]{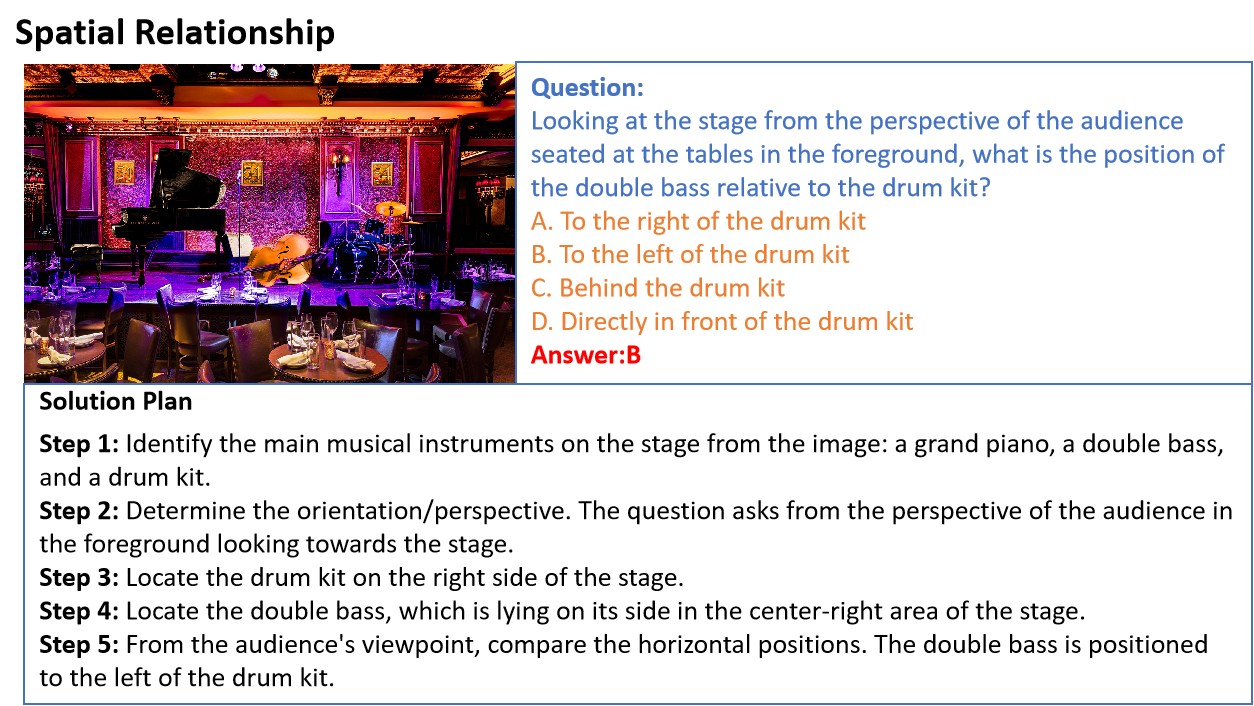}
\end{center}
\begin{center}
  \includegraphics[width=\linewidth,height=0.45\textheight]{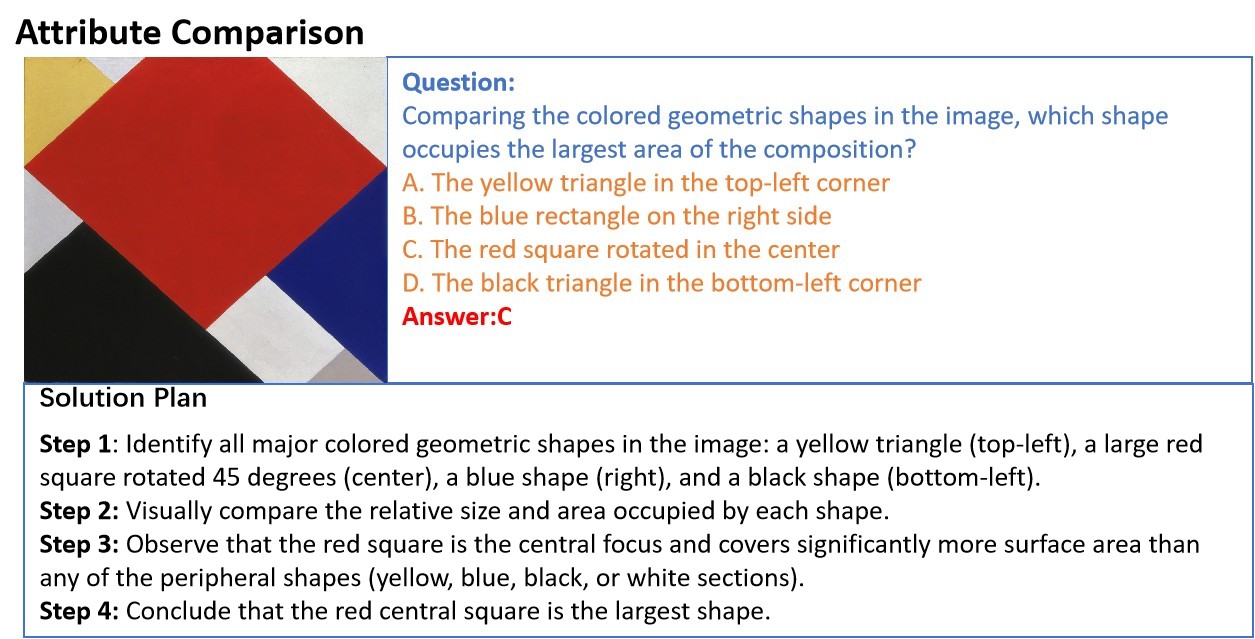}

  \includegraphics[width=\linewidth,height=0.45\textheight]{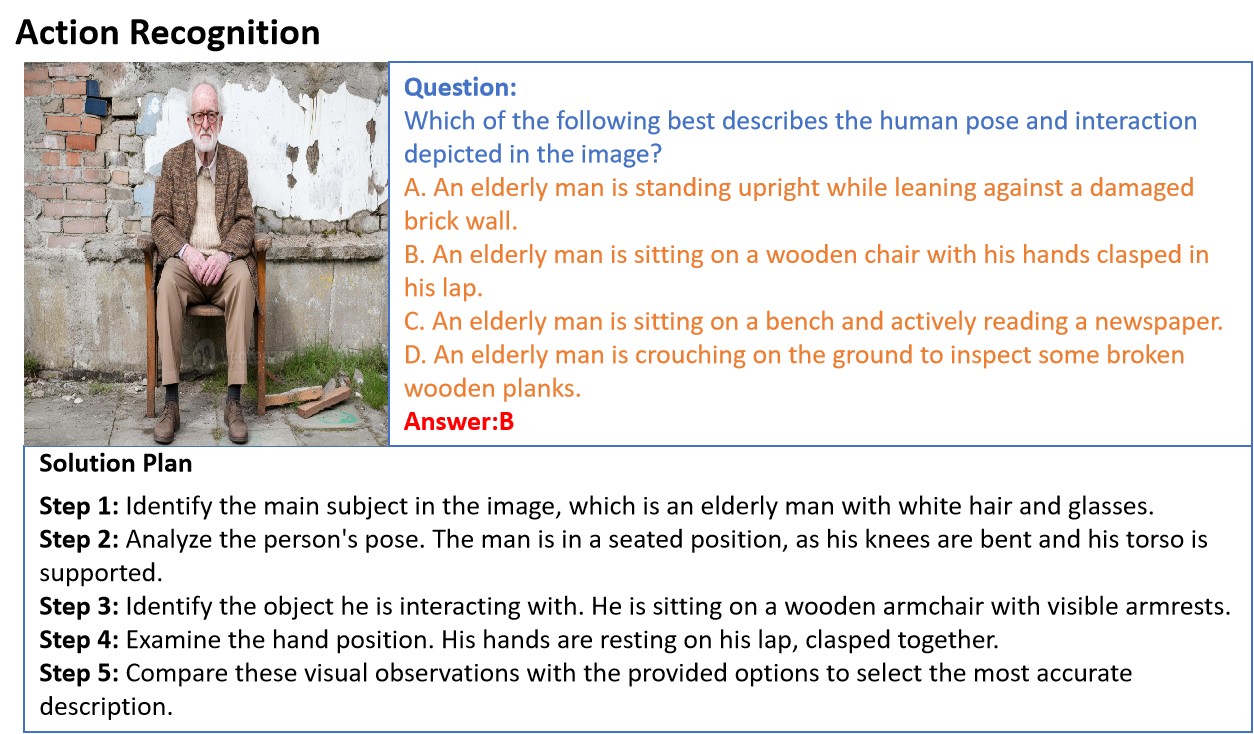}
\end{center}

\begin{center}
  \includegraphics[width=\linewidth,height=0.45\textheight]{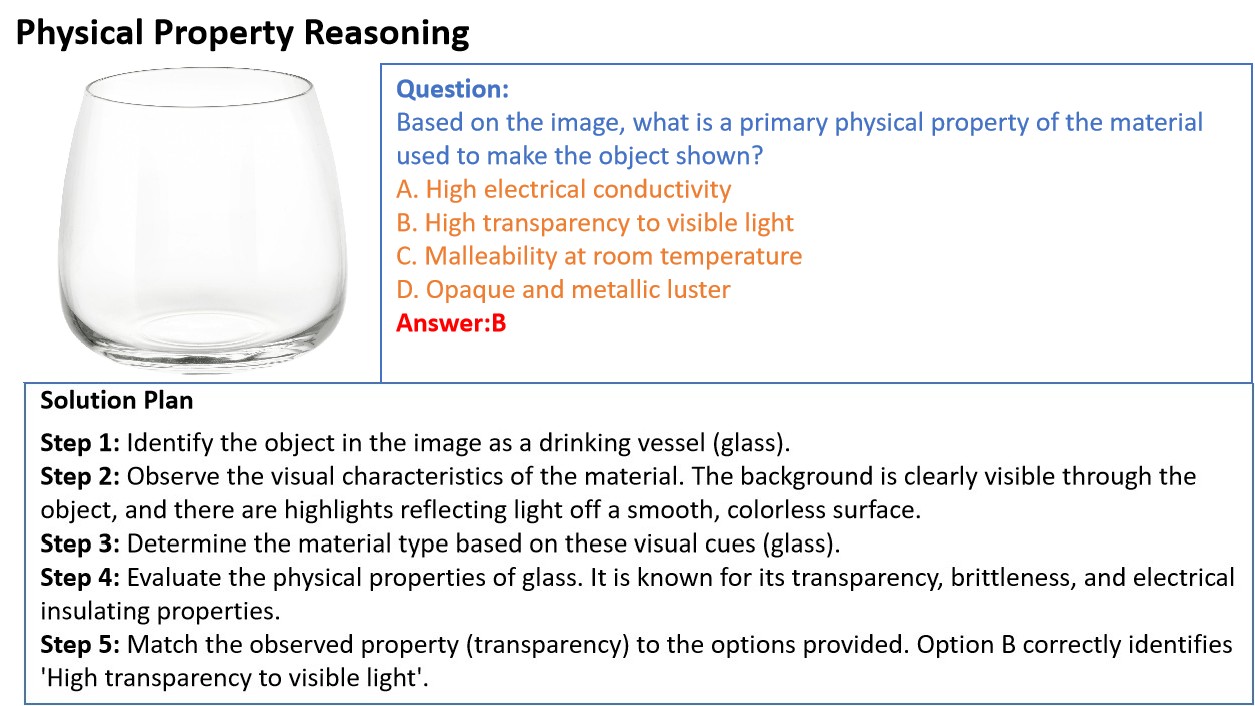}

  \includegraphics[width=\linewidth,height=0.45\textheight]{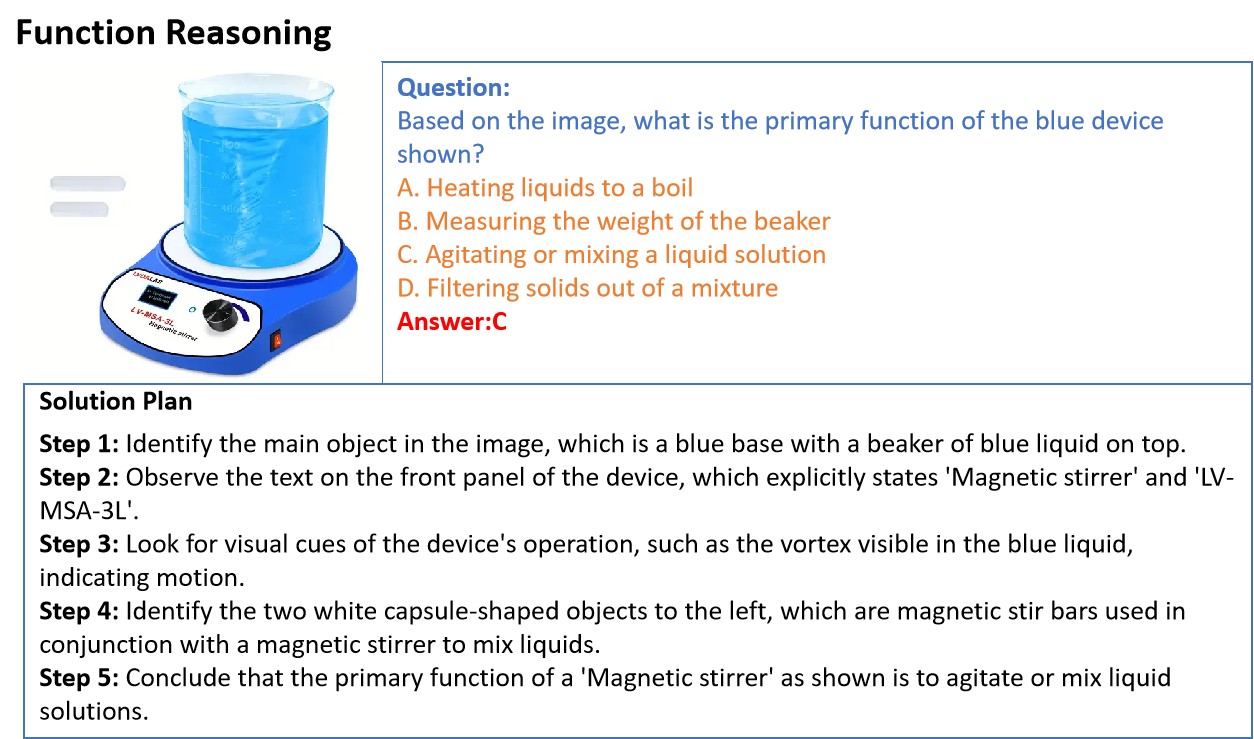}
\end{center}

\begin{center}
  \includegraphics[width=\linewidth,height=0.45\textheight]{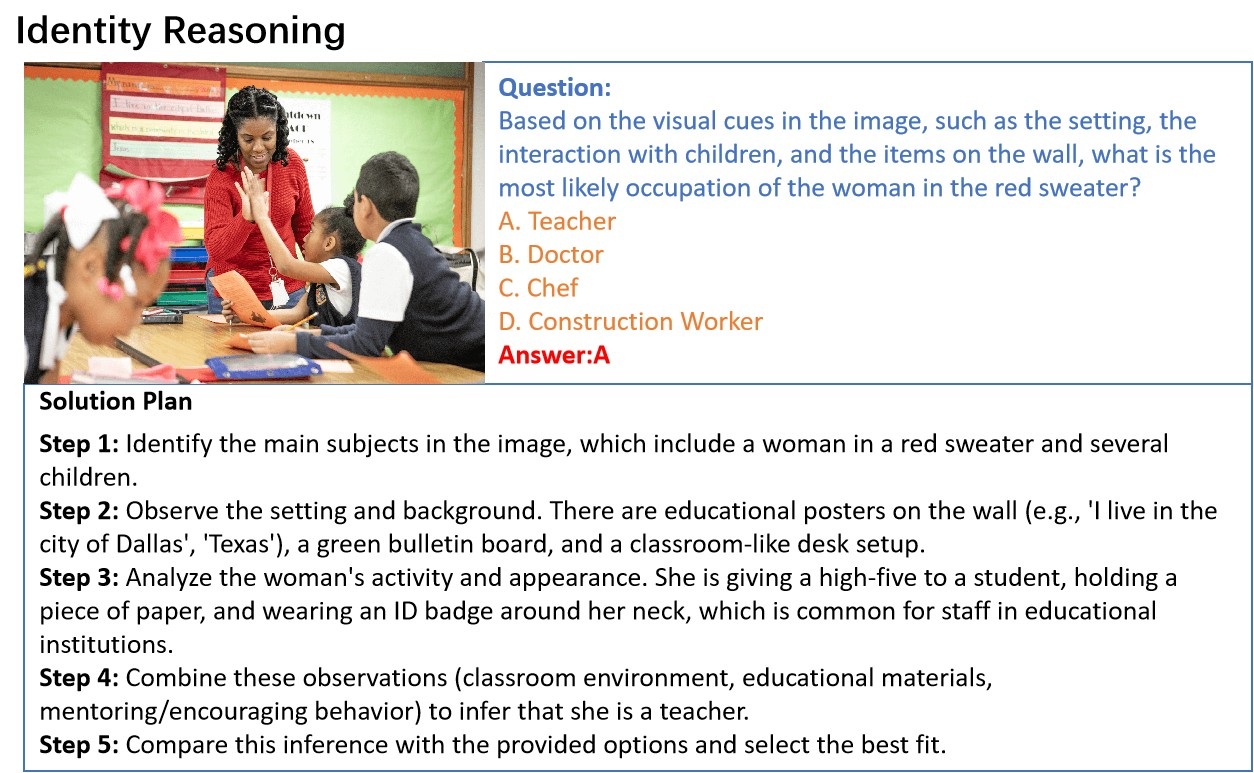}

  \includegraphics[width=\linewidth,height=0.45\textheight]{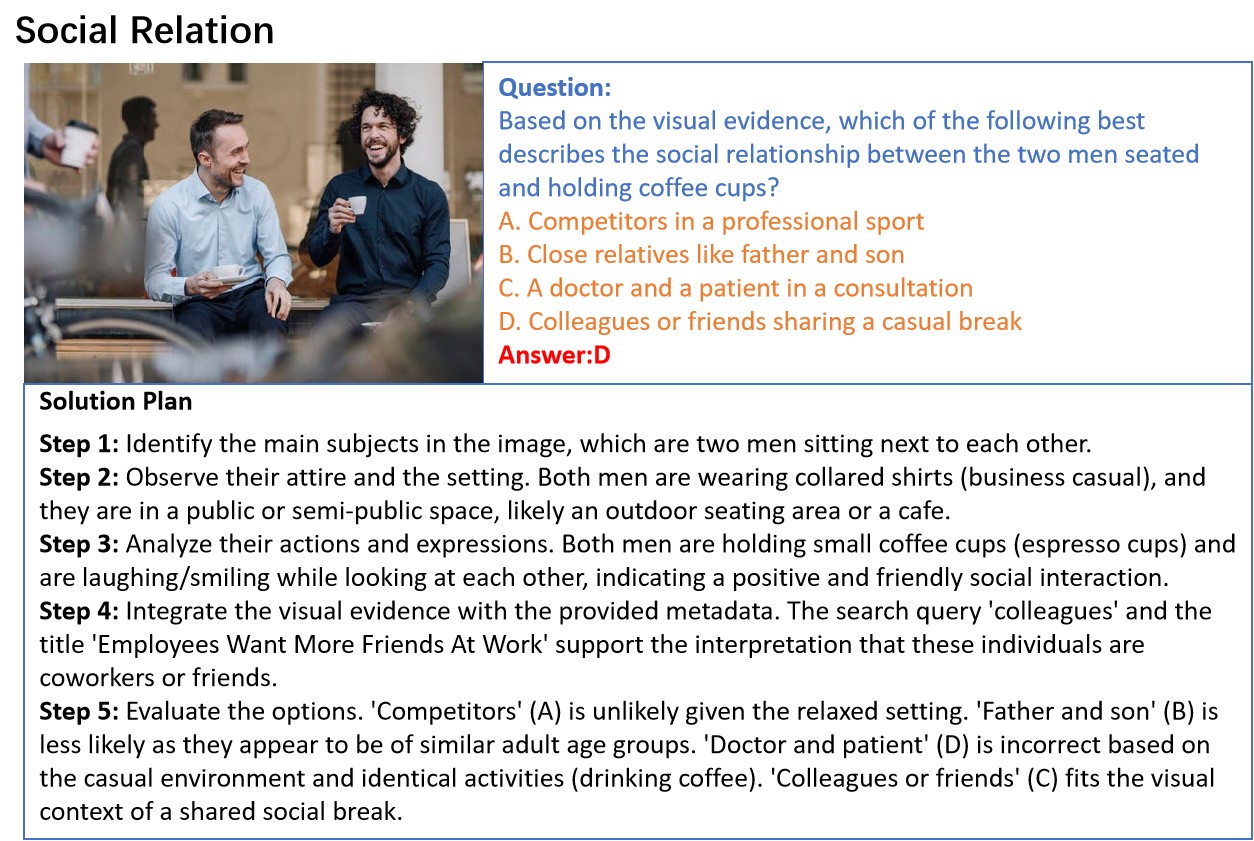}
\end{center}
\begin{center}
  \includegraphics[width=\linewidth,height=0.47\textheight]{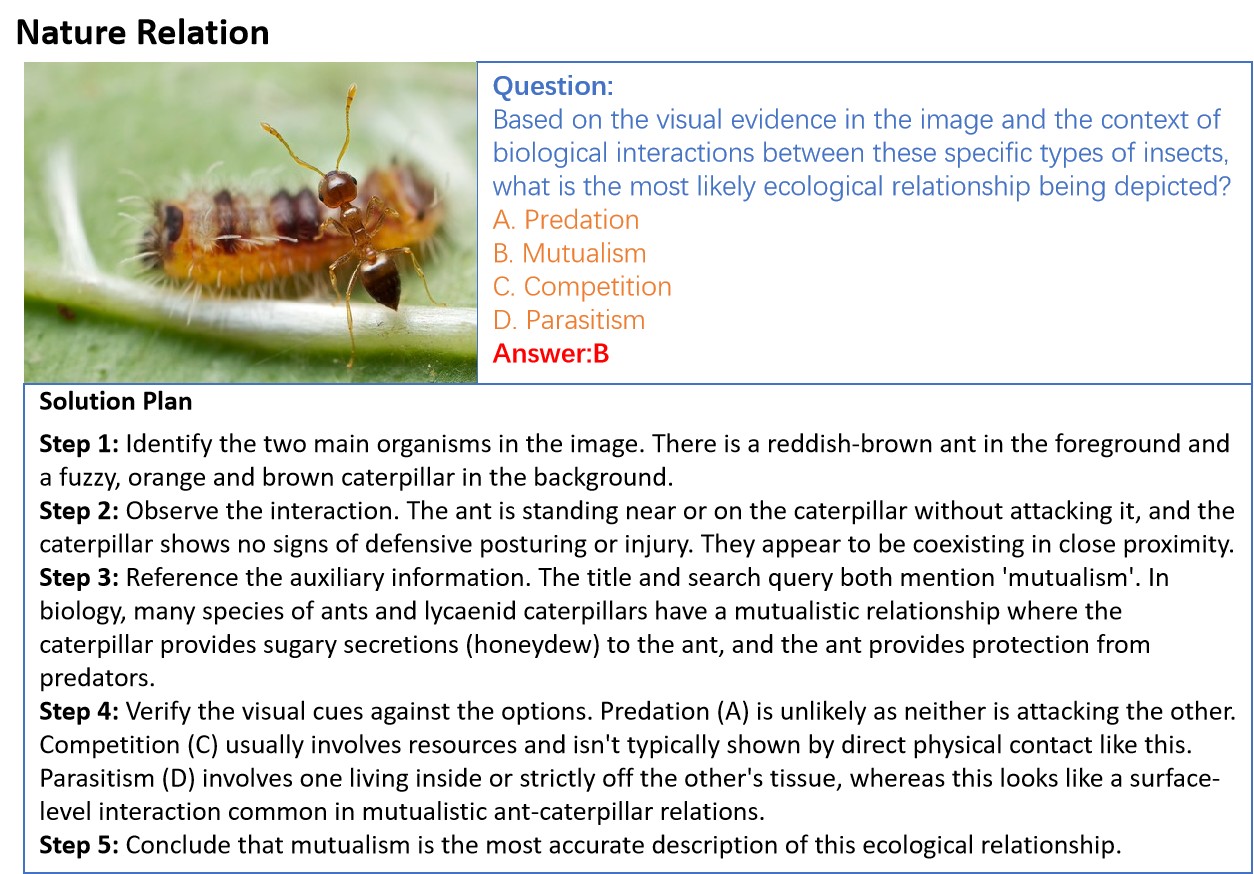}

  \includegraphics[width=\linewidth,height=0.45\textheight]{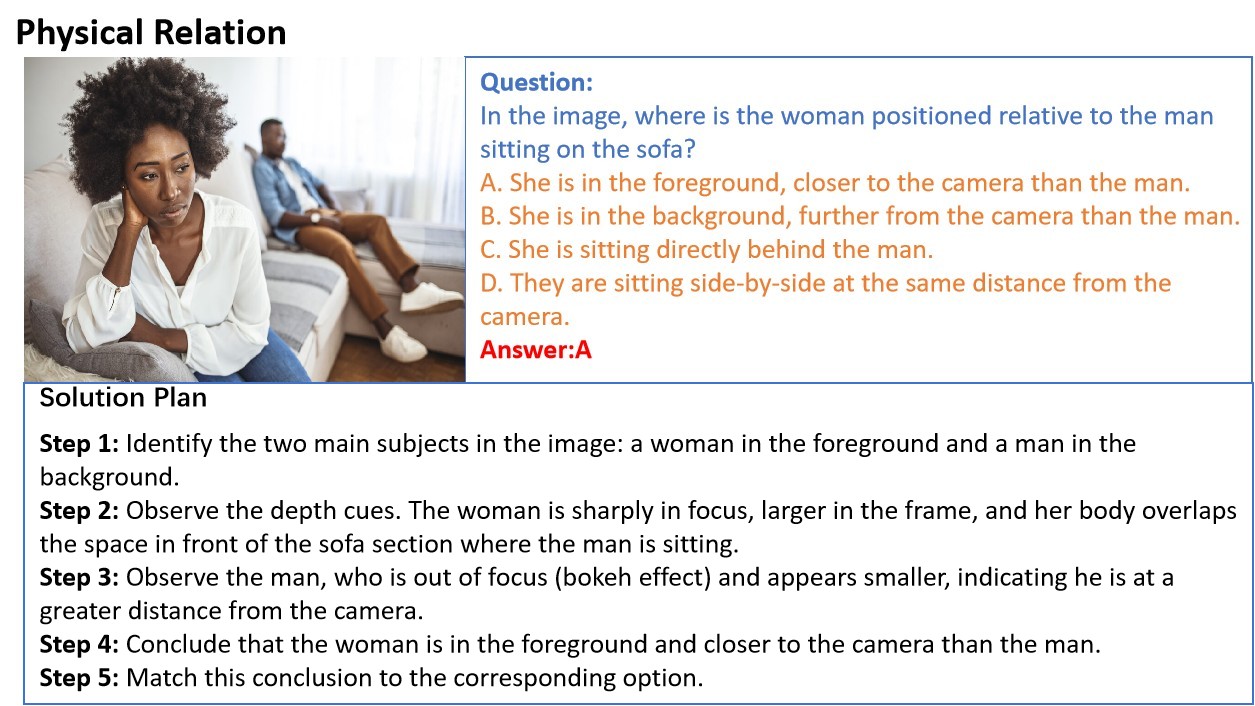}
\end{center}

\begin{center}
  \includegraphics[width=\linewidth,height=0.45\textheight]{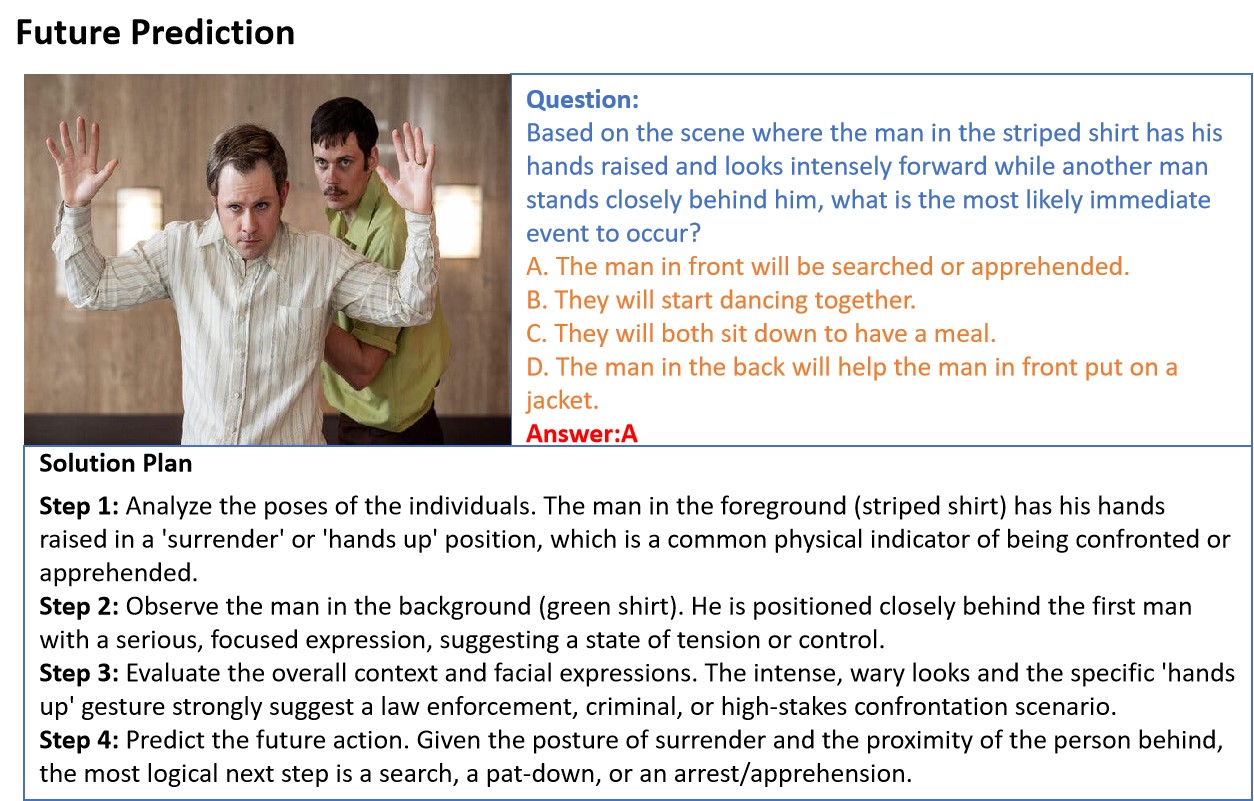}

  \includegraphics[width=\linewidth,height=0.45\textheight]{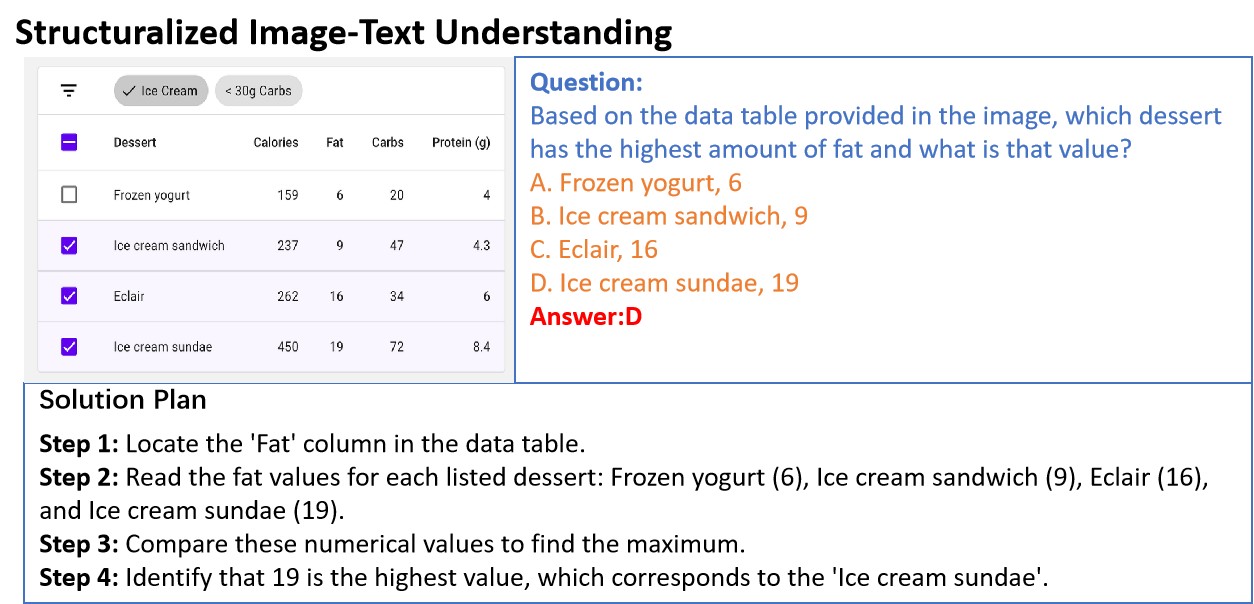}
\end{center}

\end{document}